\pdfoutput=1
\documentclass[11pt]{article}

\usepackage[final]{acl}
\usepackage{amsmath}
\usepackage{times}
\usepackage{latexsym}
\usepackage{amssymb}
\usepackage{multicol}
 \usepackage{booktabs}
 \usepackage{graphicx}
 \newtheorem{thm}{\bf Theorem}
\newtheorem{lemma}{Lemma}

\usepackage[T1]{fontenc}
\usepackage[utf8]{inputenc}

\usepackage{microtype}

\usepackage{inconsolata}
\usepackage{subcaption}
\usepackage{color, colortbl}
\usepackage{algorithm}
\usepackage{algpseudocode}
\usepackage{graphicx}
\definecolor{darkgreen}{rgb}{0.0, 0.5, 0.0}
\definecolor{light-gray}{gray}{0.95}

\newcommand\torevise[1]{\textcolor{black}{#1}}

\usepackage{soul}

\title{daDPO: Distribution-Aware DPO for Distilling Conversational Abilities}
\author{Zhengze Zhang\textsuperscript{1}\textsuperscript{2}\footnotemark[1]
\qquad Shiqi Wang\textsuperscript{1}\textsuperscript{2}\footnotemark[1] 
\qquad Yiqun Shen\textsuperscript{1}\textsuperscript{2} 
\qquad Simin Guo\textsuperscript{3} \\\bf Dahua Lin\textsuperscript{4} Xiaoliang Wang\textsuperscript{1}\textsuperscript{2} Cam Tu Nguyen\textsuperscript{1}\textsuperscript{2}\footnotemark[2] Fei Tan\textsuperscript{5}\footnotemark[2]  \qquad\\
         \textsuperscript{1}State Key Laboratory for Novel Software Technology, Nanjing University \\ \textsuperscript{2}School of Artificial Intelligence, Nanjing University\\
         \textsuperscript{3}University of Chicago
         \textsuperscript{4}The Chinese University of Hong Kong\\
         \textsuperscript{5}East China Normal University\\
         \texttt{\{zzzhang,wangsky\}@smail.nju.edu.cn}\\
         \texttt{ncamtu@nju.edu.cn}\\
         \texttt{tanfei2007@gmail.com}
         }

\begin{document}
\maketitle
\renewcommand{\thefootnote}{\fnsymbol{footnote}}
\footnotetext[1]{Zhang and Wang contributed equally to this work: Zhang leads experiments and refined the methodology, while Wang originates the idea and leads the theoretical proofs. All authors contributed to the writing, discussion, and overall development of the paper.}
\footnotetext[2]{Corresponding authors}

\begin{abstract}

Large language models (LLMs) have demonstrated exceptional performance across various applications, but their conversational abilities decline sharply as model size decreases, presenting a barrier to their deployment in resource-constrained environments. Knowledge distillation with Direct Preference Optimization (dDPO) has emerged as a promising approach to enhancing the conversational abilities of smaller models using a larger teacher model. However, current methods primarily focus on ``black-box'' KD, which only uses the teacher’s responses, overlooking the output distribution offered by the teacher. This paper addresses this gap by introducing \textit{daDPO} (Distribution-Aware DPO), a unified method for preference optimization and distribution-based distillation. We provide rigorous theoretical analysis and empirical validation, showing that \textit{daDPO} outperforms existing methods in restoring performance for pruned models and enhancing smaller LLM models. Notably, in in-domain evaluation, our method enables a 20\% pruned Vicuna1.5-7B to achieve near-teacher performance (-7.3\% preference rate compared to that of dDPO's -31\%), and allows Qwen2.5-1.5B to occasionally outperform its 7B teacher model (14.0\% win rate).
%Our approach enables 1.5B and 0.5B models to achieve competitive performance with their 7B teacher while maintaining capabilities on downstream tasks.

%Large language models (LLMs) demonstrate significantly better conversational abilities than their smaller counterparts, making effective knowledge transfer between them a critical challenge. While traditional distillation methods focus on matching predicted outputs, they often overlook the rich distributional information encoded in teacher models. With the increasing availability of open-source LLMs, we propose a novel, theoretically-grounded approach that integrates teacher model's distributional information into Direct Preference Optimization (DPO) for enhanced student model performance. Our method preserves DPO's simplicity and theoretical guarantees while successfully incorporating teacher distribution information. Experimental results demonstrate consistent improvements across both in-domain(pair comparison judged by GPT4) and out-of-domain evaluations(Alpaca-eval and MTbench), with no additional labeling costs. Notably, in in-domain evaluation, our method enables a 20\% pruned Vicuna-7B to achieve near-teacher performance (-7.3\% preference rate), and allows Qwen2.5-1.5b to occasionally outperform its 7b teacher model (14\% preference rate).
\end{abstract}

\section{Introduction}
\begin{figure}[t]
\centering
  \includegraphics[width=0.98\linewidth]{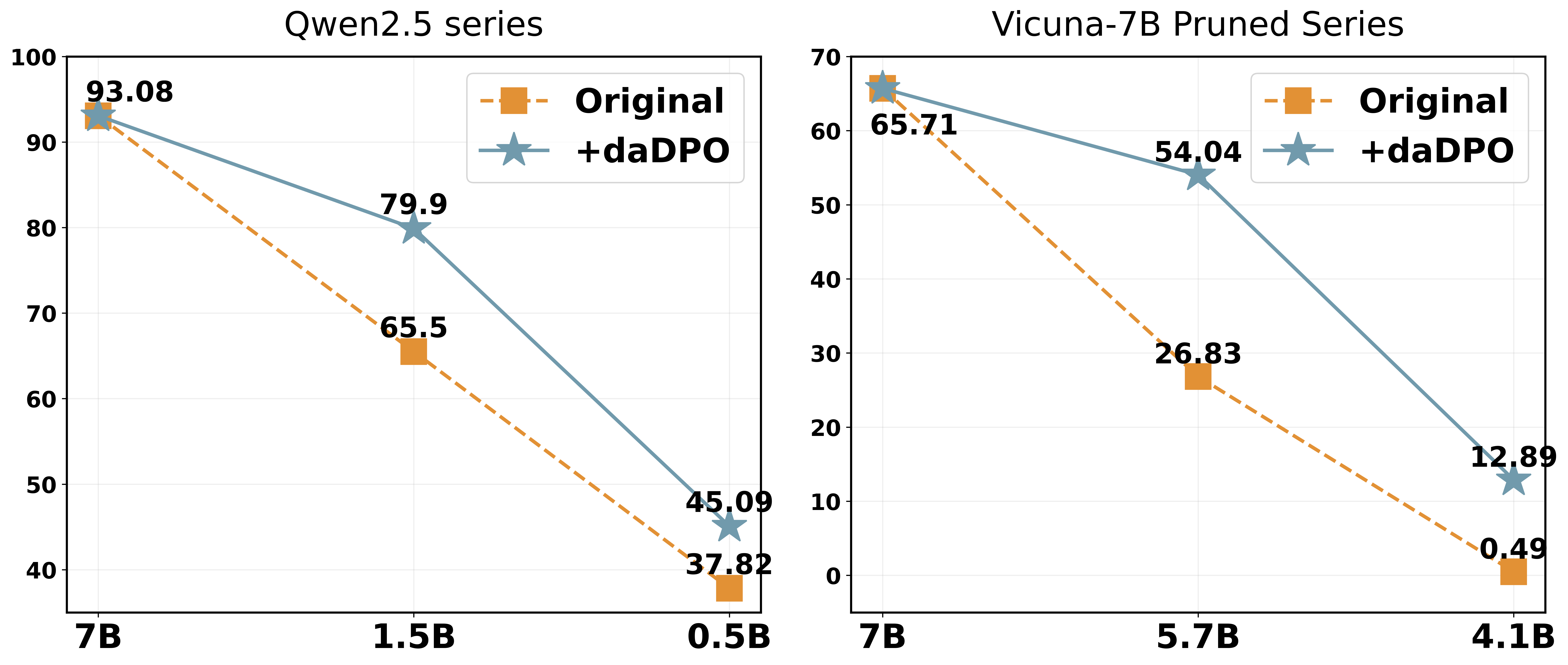}
  \caption {Conversation ability (AlpacaEval scores) drops significantly when the model gets smaller.}
  \label{fig:alpaca-eval}
\end{figure}

Large language models (LLMs) have demonstrated remarkable success across various applications \cite{gpt4-2023,claude,llama2,qwen2.5,liu2024deepseekv2,tanfei2_dch_scalinglaw,sun2025mitigating,sun2025parrot}. However, a key challenge in their deployment is the sharp decline in conversational ability as model size decreases. As shown in Figure~\ref{fig:alpaca-eval}, smaller models experience a sharp performance drop. For example, Qwen2.5-0.5B achieves an AlpacaEval score of only 37.8\%, compared to 93.8\% for Qwen2.5-7B. Since conversational ability is essential for effective human interaction across tasks and domains—requiring both intent understanding and response generation aligned with human values—this limitation poses a major barrier to the adoption of LLMs in resource-constrained environments.

Knowledge Distillation (KD) with DPO (Direct Preference Optimization) \cite{zephyr} has recently emerged as a promising approach to enhance the conversational abilities of a student model (small model) using a teacher model (large model). Specifically, responses generated by the teacher can be treated as the preferred (winning) responses, while those from the student serve as the less preferred (losing) responses, forming a preference dataset. The student model is then aligned using DPO \cite{dpo} based on the dataset. Prior work \cite{zephyr, qdpo} has demonstrated that DPO-based distillation can significantly improve the conversational performance of weaker models.  A key advantage of this approach is the automatic generation of a preference dataset, reducing the reliance on large-scale human-annotated preference data. However, existing studies primarily focus on \textit{black-box} KD, leveraging only the teacher’s responses while overlooking the rich information embedded in the teacher’s probability distribution or internal representations.

The increasing availability of powerful open-source LLMs presents a unique opportunity to explore white-box distillation with broad practical implications.  For instance, small models of various sizes can be developed through pruning or quantization of LLM, followed by knowledge distillation \cite{qdpo,prune_kd}. By doing so, one can avoid the need to build small models by from scratch, significantly reducing development costs. \torevise{White-box distillation is potentially useful as shown in knowledge distillation literature for text classification \cite{sanh2019distilbert}, image classification \cite{hinton2015distilling}, LLM \cite{minillm_kl}, however, there is still a research gap in leveraging white-box distillation to enhance the conversational ability of small language models.} Specifically, existing studies primarily focus on either supervised knowledge distillation \cite{prune_kd} or online reinforcement learning \cite{onpolicy_distillation, minillm_kl}. The former is suboptimal for aligning conversational ability \cite{instructgpt2022}, whereas the latter—though adaptable for alignment (e.g., by training the reward model) —is more computationally expensive and challenging to optimize compared to DPO \cite{dpo}. 

To address this gap, we focus on the critical research question: \textit{“How can we adapt DPO-based distillation to incorporate the teacher’s distribution for enhancing the student's conversational ability?”} Our preliminary results suggest that directly combining the Kullback–Leibler (KL) divergence between the student's and teacher's distributions—a common metric in knowledge distillation—into the final DPO loss does not yield optimal results. Instead, we begin by incorporating the KL metric into the RL objective. Our goal is to maximize the reward that favors the teacher's response over the student's response, while also regulating the student's distribution (over its own response) more closely with that of the teacher. We then derive a new DPO-style loss for distribution-aware DPO (termed \textit{daDPO}) from this RL objective. Additionally, we explore further considerations, such as the role of the reference model in DPO, the size of the teacher's model, and the possibility to distill across LLM families.

Our contributions: 1) \textbf{Novel Integration Framework}: \textit{daDPO} offers a unified objective that simultaneously facilitates preference optimization and ``white-box'' KD; 2) \textbf{Theoretical Analysis}: We provide theoretical analysis and show how we should balance between training stability and the ``white-box'' KD objective. Finally, 3) \textbf{Empirical Validation}: we evaluate the effectiveness of \textit{daDPO} with various scenarios and empirical analysis.

\section{Related Work}
\paragraph{Language Model Alignment}
Aligning language models to human preferences is crucial for their conversational ability. Reinforcement Learning with Human Feedback (RLHF) has been introduced as an effective tool for LLM alignment \cite{instructgpt2022,summarize2020}. To mitigate the cost and complexity of RLHF, numerous alternative methods have been proposed, such as SLic-HF \cite{SLiC-HF2023}, DPO \cite{dpo}, and GRPO \cite{liu2024deepseekv2}. Among these, DPO and its variants \cite{ipo,ethayarajh2024kto,meng2024simpo,emnlp_wsq} have gained prominence due to their strong theoretical foundations and robust performance. Despite these advancements, the conversational ability of smaller models remains limited (see Figure~\ref{fig:alpaca-eval}). In this work, we focus on distilling the conversational capabilities of a large LM into smaller models and propose an effective distillation method based on DPO.

% \todo{Despite the success of alignment, the conversational ability of small -chat models still not match that of larger models as shown in Figure 1. Recent studies have shown that distillation can help develop small models by pruning and recovering, without the need to train from scratch with large training data \cite{prune_kd}. As a result, we are interested in developing a better distillation strategy for enhancing small models capability.}
% The reference policy in the RLHF plays an important role in smoothing the training and making the performance more robust. \citet{f_divergence_mini} proposes to use different kinds of KL divergence in the DPO formulation and finds that loss with forward KL achieves better diversity than vanilla reverse KL DPO. \citet{sjtu_policymodel} proposes to replace the reference policy in DPO loss with a stronger model and shows that such a replacement can sometimes benefit. \citet{onpolicy_distillation} directly replaces the KL term in the vanilla RL loss with distillation KL between student and teacher LLM and still leverages PPO \cite{ppo} to optimize the loss. Unlike them, we propose to combine two KL terms, the reference model KL constraint and the teacher model KL distillation, together to better improve the student LLM's conversational ability in a distillation paradigm.

\paragraph{Language Model Distillation}
Knowledge Distillation (KD) is a technique for training a smaller model (the student) by leveraging knowledge from a larger model (the teacher). In the era of LLMs, KD is widely used to reduce the computational demands of large models by developing smaller LLMs under the guidance of a larger teacher model.  

Recent KD methods for LLMs can be broadly categorized into two types: \textit{black-box KD}, where only the teacher-generated texts are accessible, and \textit{white-box KD}, where the teacher model's output distribution or intermediate hidden states are also available. Most existing KD approaches for LLMs fall under black-box KD, including \cite{baize, alpaca, vicuna, zephyr, qdpo,tanfei1_sft, smallbutfunny}. White-box KD methods have been comparably less explored \cite{minillm_kl, onpolicy_distillation}.  

Our work also focuses on the white-box KD setting. However, instead of using online RL like \cite{minillm_kl, onpolicy_distillation}, we aim to enhance conversational capabilities through a simple yet effective method based on the DPO framework.

\section{Preliminaries}\label{sec:baseline}

%\subsection{Problem Definition}

This paper investigates the problem of enhancing\textbf{ a student model $\mathbf{\pi_{st}}$} under the guidance of \textbf{a teacher model $\mathbf{\pi_{te}}$}. Unlike \cite{zephyr, qdpo}, we consider the white-box setting, where the teacher model's distribution, $\pi_{te}$, is fully accessible. Specifically, for any given context $x$ and token $t$ in the vocabulary space, we can directly obtain the probability $\pi_{te}(t|x)$ from the teacher model.  

% \begin{itemize}
%     \item One teacher model only: In order to simplify the setting, we assume only one teacher model $\pi_{te}$ is considered.
%     \item White-box Assumption: We assume the teacher model's distribution $\pi_{te}$ is fully accessible. Specifically, for any context $x$ and token $t$ in the vocabulary space, we can obtain the probability $\pi_{te}(t|x)$ directly from the model. 
%     \item No additional information: We assume no additional information from any third party excluding the teacher model. It means that we could NOT use labels from humans or stronger model like GPT4.
% \end{itemize}

\subsection{Supervised Knowledge Distillation}\label{sec:sft}

\paragraph{Supervised Finetuning (SFT)} Given a set $\mathcal{C}_{sft} = \{(x_1, y_1), \ldots, (x_N, y_N)\}$, where (x,y) are respectively the prompt and the ground-truth output, the SFT loss is calculated as follows:
\begin{align}
  \pi_{SFT}= \arg\max_\pi \mathbb{E}_{(x,y)\sim\mathcal{C}_{sft}} \log\pi(y|x) \label{eq:sft-loss}
\end{align}
\noindent Alternatively, one can treat the teacher-generated outputs as ground truth and perform distillation by applying the SFT loss to these outputs. We refer to this as $dSFT$ (Distillation by Supervised Fine-Tuning) and denote the resulting model as $\pi_{dSFT}$.

\paragraph{Supervised KD} Given the white-boxed assumption, supervised KD is widely used to distill information from teacher models \cite{minillm_kl,prune_kd}. Specifically, given a dataset $\mathcal{C}_{skd}=\{(x,y^t)\}$, where $y^t$ is the teacher model's output for the input $x$, the following KL-divergence loss is minimized:

\begin{align}
\ell_{KL}=&\mathbb{E}_{(x,y^t)\sim\mathcal{C}^{skd}}\mathcal{D}_{KL}[\pi_\theta||\pi_{te}]  \label{kl-loss}\\
=-\mathbb{E}_{x,y^t}&\left[\frac{1}{L_y}\sum_{b=1}^{L_y}\pi_\theta(\cdot|y^t_{\le n},x)\log\frac{\pi_\theta(\cdot|y^t_{\le n},x)}{\pi_{te}(\cdot|y^t_{\le n},x)}\right] \nonumber
\end{align}

\noindent Typically, this token-level KL is used along with the SFT loss or pretraining loss \cite{minillm_kl}.

\subsection{DPO-based Distillation}
% DPO distillation is recently proposed to boost the performance of the student LLM, especially for its conversational ability. Specifically, let $\mathcal{X}=[x_1,x_2,\dots,x_N]$ be a set of prompts. DPO distillation is typically performed via two stages, where the first stage is dSFT and the second one is DPO. In dSFT stage, for each $x_i \in \mathcal{X}$, we sample a response $y_i \sim \pi_{\text{te}}(\cdot \mid x_i)$ from the teacher model. At the first stage, we perform dSFT to obtain $\pi_{dSFT}$ based on teacher-generated outputs as aforcementioned. In the DPO stage, for each $x_i \in \mathcal{X}$, we sample response $y_i^t \sim \pi_{\text{te}}(\cdot \mid x_i)$ from the teacher model and $y_i^s \sim \pi_{dSFT}(\cdot\mid x_i)$ from the (dSFT) student model. A ranking dataset $\mathcal{C}_{dpo} = \{(x_1, y_1^t,y_1^s), \ldots, (x_N, y_N^t,y_N^s)\}$ is then constructed for DPO.

% It should be noted that a typical DPO dataset is a ranking dataset $\mathcal{D}=\{(x,y_w,y_l)\}$ where $y_w$ and $y_l$ stand for the winning and losing response respectively.  Here, $\mathcal{C}_{dpo}$ is used as a DPO dataset by always treating $y^s,y^t$ as losing response $y_l$ and winning response $y_w$ respectively. Such preference of $y_t$ over $y_s$ is ensured as long as the teacher model $\pi_{te}$ is the ideal optimal policy according to a previous paper \cite{qdpo}.  The DPO \cite{dpo} loss is then calculated and optimized as follows:
DPO-based distillation (dDPO) has recently been introduced to enhance the performance of small LLMs, particularly in terms of conversational ability. It typically involves two stages: dSFT (Distillation by Supervised Fine-Tuning) followed by DPO (Direct Preference Optimization). Specifically, given a set of prompts \(\mathcal{X} = [x_1, x_2, \dots, x_N]\), for each \(x_i \in \mathcal{X}\), we sample a response \(y^t_i \sim \pi_{\text{te}}(x_i)\) from the teacher model and another response \(y_i^s \sim \pi_{st}(x_i)\) from the student model.  

In the dSFT stage, the student model is fine-tuned on these teacher-generated outputs to obtain \(\pi_{dSFT}\), as described in Section \ref{sec:sft}. In the DPO stage, we treat \textit{\textbf{\(y^t\) as the winning response}} and \textit{\textbf{\(y^s\) as the losing response}}, forming a preference dataset \(\mathcal{C}_{dpo} = \{(x_1, y_1^t, y_1^s), \dots, (x_N, y_N^t, y_N^s)\}\). The preference for \(y^t\) over \(y^s\) is justified under the assumption that the teacher model \(\pi_{te}\) represents an optimal policy, as outlined in prior work \cite{qdpo}. Finally, distillation is conducted by optimizing the DPO loss \cite{dpo}:  
\begin{align} 
\ell_{dDPO}=-\mathop{\mathbb{E}}\bigg[\log\sigma\bigg(&\beta\frac{\log\pi_\theta(y^{t}|x)}{\log{\pi_{ref}(y^{t}|x)}}-\nonumber\\&\beta\frac{\log\pi_\theta(y^{s}|x)}{\log{\pi_{ref}(y^{s}|x)}}\bigg)\bigg] \label{eq:dpo_loss}
\end{align}

% \begin{align} 
% \ell_{DPO}=-\mathop{\mathbb{E}}\limits_{(x,y^t,y^{st})\sim\mathcal{C}_{dpo}}\bigg[\log\sigma\bigg(&\beta\frac{\log\pi_\theta(y^{t}|x)}{\log{\pi_{ref}(y^{t}|x)}}\nonumber\\&-\beta\frac{\log\pi_\theta(y^{st}|x)}{\log{\pi_{ref}(y^{st}|x)}}\bigg)\bigg] \label{eq:dpo_loss}
% \end{align}
\noindent The expectation is computed over $(x, y^t, y^s) \sim \mathcal{C}_{dpo}$. Here, the reference policy \(\pi_{ref}\) is  set to \(\pi_{dSFT}\), while \(\pi_\theta\) represents the student model being optimized. The student model is initialized with \(\pi_{dSFT}\).

\section{Distribution-aware DPO for Distillation}\label{sec:dadpo}
\subsection{A Direct Approach: DPO with Reference Replacement}\label{sec:replacement}
\torevise{Similar in some previous studies \cite{li2024direct,sjtu_policymodel}}, our initial attempt was to directly incorporate the KL-loss from Eq. \ref{kl-loss} into the DPO loss from Eq. \ref{eq:dpo_loss}. We refer to this approach as \textit{dDPO+KL}. This method, however,  does not preserve the theoretical guarantees offered by \textit{DPO}. As an alternative, we start with the Reinforcement Learning (RL) objective, which, as shown by \citet{dpo}, yields the same optimal solution \(\pi^*\) as optimizing the DPO loss (Eq. \ref{eq:dpo_loss}):
\begin{align}
    \max \mathbb{E}_{x\sim\mathcal{D},y\sim\pi_\theta}[r(x,y)]-\beta\mathcal{D}_{KL}[\pi_\theta||\pi_{ref}] \label{rl-loss}
\end{align}
where \(r(x, y)\) represents the reward model that favors the teacher's response. In the context of DPO, \(r(x, y)\) corresponds to the implicit reward model \cite{dpo}. To integrate the teacher's distribution, we replace the reference model in Eq. \ref{rl-loss} with the teacher model, resulting in the new RL loss as follows:
\begin{align}
    \max \mathbb{E}_{x\sim\mathcal{D},y\sim\pi_\theta}[r(x,y)]-\beta\mathcal{D}_{KL}[\pi_\theta||\pi_{te}] \label{rl-loss-rdpo}
\end{align}

This loss can be broken down into two parts: the first part optimizes the reward that prioritizes the teacher response over the student response (the alignment objective), while the second part distills knowledge based on the teacher model's distribution. Simple algebra demonstrates that this approach effectively replaces the reference model in Eq. \ref{eq:dpo_loss} with the teacher model's distribution.

\begin{lemma}
The loss in Eq. \ref{rl-loss-rdpo} has the same optimal policy \(\pi^*\) as the optimization of the following loss:
\begin{align}
    \ell_{rDPO}=-\mathop{\mathbb{E}}\bigg[\log\sigma\bigg(&\beta\frac{\log\pi_\theta(y^t|x)}{\log{\pi_{te}(y^t|x)}}-\nonumber\\&\beta\frac{\log\pi_\theta(y^s|x)}{\log{\pi_{te}(y^s|x)}}\bigg)\bigg] \label{rdpo-loss}
\end{align}
\end{lemma}

\noindent where the expectation is computed over \((x, y^t, y^s) \sim \mathcal{C}_{dpo}\), similar to DPO. This loss (\textit{rDPO}), however, discards the reference policy constraint from the original RL loss. This reference model helps prevent large policy updates and ensure stable learning \cite{ppo,dpo}. %Thus, we believe such replacement is suboptimal.

% One may notice that both RL loss in eq.\ref{rl-loss} and KL loss in eq.\ref{kl-loss} leverage reverse KL. Then is it possible to combine these two together to include more teacher model distribution information during DPO training? In this section, we first introduce a direct but suboptimal solution: reference replacement. 

% If we directly replace the reference policy in loss \ref{eq:dpo_loss} to teacher model $\pi_{te}$ as follows:
% \begin{align}
%     \ell_{Rdpo}=-\mathop{\mathbb{E}}\limits_{(x,y_w,y_l)\sim\mathcal{D}}\bigg[\log\sigma\bigg(&\beta\frac{\log\pi_\theta(y_w|x)}{\log{\pi_{te}(y_w|x)}}\nonumber\\&-\beta\frac{\log\pi_\theta(y_l|x)}{\log{\pi_{te}(y_l|x)}}\bigg)\bigg] \label{rdpo-loss}
% \end{align}
% And replace the reference policy in the proof of DPO loss, we would have:

% \begin{lemma}
%     The loss \ref{rdpo-loss} has the same optimal policy $\pi^*$ as:
% \begin{align}
%     \max \mathbb{E}_{x\sim\mathcal{D},y\sim\pi_\theta(x)}[r(x,y)]-\beta*\mathcal{D}_{KL}[\pi_\theta||\pi_{te}] \label{rl-loss}
% \end{align}
% \end{lemma}

\subsection{The Joint Force of Two Distributions}
%Earlier attempts like reference replacement showed promise but highlighted a key challenge: while incorporating teacher distribution information is valuable, we can't simply discard the reference policy constraint that makes DPO training stable. This raises an important question: Can we capture the best of both worlds?

The limitation of $rDPO$ motivates us to propose Distillation-Aware DPO ($daDPO$), which reintroduces the reference distribution into the \textit{rDPO} equation. Specifically, we seek to optimize the following RL objective.
\begin{align}
    \max&\mathop{\mathbb{E}}\limits_{x,y\sim \pi_\theta}\left[r(x,y)\right]\nonumber-\beta_1D_{KL}[\pi_\theta(y|x)||\pi_{ref}(y|x)]\\&-\beta_2D_{KL}[\pi_\theta(y|x)||\pi_{te}(y|x)] \label{original_loss}
    % \max&\mathop{\mathbb{E}}\limits_{x,y\sim \pi_\theta}r(x,y)\nonumber-\beta_1D_{KL}[\pi_\theta(y|x)||\pi_{ref}(y|x)]\\&-\beta_2D_{KL}[\pi_\theta(y|x)||\pi_{te}(y|x)] \label{original_loss}
\end{align}
where $\pi_{te}$ represents the teacher LLM and $\pi_{ref}$ for the reference LLM (usually the student LLM).
\begin{thm}
The optimal solution of the objective defined in Eq.\ref{original_loss} is:
\begin{align}
    \pi_\theta^*(y|x)=&\frac{1}{Z(x)}\pi_{ref}(y|x)^\frac{\beta_1}{\beta_1+\beta_2}\pi_{te}(y|x)^{\frac{\beta_2}{\beta_1+\beta_2}}\nonumber\\&\times\exp(\frac{1}{\beta_1+\beta_2}r(x,y)) 
\end{align}
\noindent where $Z(x)=\sum_{y}\pi_{ref}(y|x)^\frac{\beta_1}{\beta_1+\beta_2}\times\pi_{te}(y|x)^{\frac{\beta_2}{\beta_1+\beta_2}}\exp(\frac{1}{\beta_1+\beta_2}r(x,y))$ is a constant that depends solely on the input $x$.
\label{thm:optimal_solution}
\end{thm}
From Theorem 1, we can derive the following relationship between the reward model and the optimal model $\pi^*$:
\begin{align}
    r(x,y)=&\log\frac{\pi_\theta^*(y|x)^{\beta_1+\beta_2}}{\pi_{ref}(y|x)^{\beta_1}\pi_{te}(y|x)^{{\beta_2}} }\nonumber\\&+(\beta_1+\beta_2) \log Z(x)
\end{align}
Since \(\log Z(x)\) is independent of \(y\), under the Bradley-Terry model, the optimal policy $\pi^*$ satisfies the preference model $p(y^t\succ y^s|x)$, which not only depends on the optimal policy \(\pi^*\), the reference policy \(\pi_{ref}\) like DPO but also the teacher distribution \(\pi_{te}\). Please refer to the Appendix Eq. \ref{eq:long-p} for more details.
% \begin{equation}
%     p^*(y_1\succ y_2)= \frac{1}{1+\exp(\log\frac{\pi_\theta^*(y_2|x)^{\beta_1+\beta_2}}{{\pi_{ref}(y_2|x)^{\beta_1}\pi_{te}(y_2|x)^{\beta_2}}}-\log\frac{\pi_\theta^*(y_1|x)^{\beta_1+\beta_2}}{{\pi_{ref}(y_1|x)^{\beta_1}\pi_{te}(y_1|x)^{\beta_2}}})}
% \end{equation}
Accordingly, the loss function becomes:
\begin{align}
    %\ell=
    -\mathop{\mathbb{E}}\Bigg[&\log\sigma\Bigg(\log\frac{\pi_\theta(y^t|x)^{\beta_1+\textcolor{blue}{\beta_2}}}{{\pi_{ref}(y^t|x)^{\beta_1}\textcolor{blue}{\pi_{te}(y^t|x)^{\beta_2}}}}\nonumber\\&-\log\frac{\pi_\theta(y^s|x)^{\beta_1+\textcolor{blue}{\beta_2}}}{{\pi_{ref}(y^s|x)^{\beta_1}\textcolor{blue}{\pi_{te}(y^s|x)^{\beta_2}}}}  \Bigg)\Bigg] \label{daDPO_loss}
\end{align}
The key difference between \textit{daDPO} and the original DPO loss (Eq. \ref{eq:dpo_loss}) is highlighted in blue. Notably, when \(\beta_2 = 0\), the loss in Eq. \ref{daDPO_loss} simplifies to the original DPO loss in Eq. \ref{eq:dpo_loss}. Likewise, when \(\beta_1 = 0\), it becomes equivalent to the rDPO loss in Eq. \ref{rdpo-loss}.  Our derivation maintains the simplicity and the theoretical analysis of DPO.

\paragraph{Gradient Analysis}
The gradient of the daDPO loss in Eq. \ref{daDPO_loss} can be derived as follows:
\begin{align}
    \nabla\ell=&-\nabla \mathop{\mathbb{E}}\Bigg[\log\sigma\Bigg(\log\frac{\pi_\theta(y^t)^{\beta_1+\beta_2}}{{\pi_{ref}(y^t)^{\beta_1}\pi_{te}(y^t)^{\beta_2}}}\nonumber\\&-\log\frac{\pi_\theta(y^s)^{\beta_1+\beta_2}}{{\pi_{ref}(y^s)^{\beta_1}\pi_{te}(y^s)^{\beta_2}}}  \Bigg)\Bigg] \nonumber \\
    =&-\nabla \mathop{\mathbb{E}}\ \sigma\bigg[(\beta_1+\beta_2)(-\delta_\theta)+
\beta_2\delta_{te}\bigg]\times\nonumber\\&\bigg[(\beta_1+\beta_2)(\nabla_{\theta}\log \pi(y^t)-\nabla_{\theta}\log\pi(y^s))\bigg] \nonumber
\end{align}
where $\delta_\theta=\log\frac{\pi_\theta(y^t)}{\pi_{ref}(y^t)}-\log\frac{\pi_\theta(y^s)}{\pi_{ref}(y^s)}$ and $\delta_{te}=\log\frac{\pi_{te}(y^t)}{\pi_{ref}(y^t)}-\log\frac{\pi_{te}(y^s)}{\pi_{ref}(y^s)}$. 

Similar to the original DPO, the gradient of the loss function increases the likelihood of the chosen response \( y^t \) while decreasing that of the rejected response \( y^s \). The gradient is scaled by a coefficient consisting of two terms: \( -(\beta_1+\beta_2)\delta_\theta \) and \( \beta_2\delta_{te} \).  

The first term, \( -(\beta_1+\beta_2)\delta_\theta \), represents how much the implicit reward model of the currently optimized policy \( \pi_{\theta} \) favors the dispreferred completions, scaled by \( \beta_1+\beta_2 \). The second term, \( \beta_2\delta_{te} \), captures how much the implicit reward model of the teacher model $\pi_{te}$ favors the preferred completions. The more strongly the teacher model prefers the chosen response over the rejected one, the larger this coefficient becomes.

\paragraph{Sentence-Level KL vs. Token-Level KL}  In both \textit{dDPO} and \textit{daDPO}, the KL divergence used in the RL objective (Eq. \ref{rl-loss}) is computed at the sentence level, as the problem is framed as a sentence-level Markov Decision Process (MDP). In contrast, KL divergence for KD is typically calculated at the token level.  Recent studies \cite{qfunction,token_dpo}, however, show that token-level DPO produces a loss function equivalent to the original sentence-level formulation. This establishes that the DPO framework remains mathematically consistent whether applied in a token-level or sentence-level MDP setting. Given this theoretical equivalence, we adopt the sentence-level MDP formulation for its mathematical simplicity. 

\paragraph{Advantages of daDPO} \textit{daDPO} offers several advantages. First, it preserves the stabilizing effect of the reference policy, preventing large policy updates. 
Second, it leverages the rich distributional information from the teacher model, enhancing distillation. 
Additionally, since \textit{daDPO} exploits sentence-level KL, it also has the advantage of being applicable across LLM families with different tokenizers, unlike most existing KD methods that rely on token-level KL. Lastly, \textit{daDPO} preserves the simplicity and theoretical guarantees of DPO. Compared to vanilla DPO, it requires only an additional teacher LLM for sampling and computing logits of sentences in the preference dataset. In other words, implementing \textit{daDPO} merely involves adding a few lines of code to the original DPO. Moreover, since the teacher model can be frozen in a gradient-free mode, the additional memory overhead and training time remain manageable.

\section{Experiments}
\subsection{Datasets and Evaluation Metrics}
Our training dataset comprises triplets \(\{(x, y^t, y^s)\}\), where \(x\) serves as the prompt. These prompts are from ShareGPT\footnote{https://huggingface.co/datasets/Aeala/ShareGPT\newline \_Vicuna\_unfiltered}, after being filtered and deduplicated. We use only the prompt from the first round of each dialogue, yielding a total of 49,839 prompts. The response \(y^t\) is generated by the teacher model \(\pi_{te}\) via greedy decoding, while \(y^s\) corresponds to the response from the student model \(\pi_{st}\). For evaluation, we conduct both in-domain assessments on ShareGPT and out-of-domain evaluations using MT-Bench and AlpacaEval.

%We mainly leverage three folds of evaluation to examine the language model's conversational ability:
\paragraph{In-domain Evaluation} To evaluate in-domain performance, we randomly sampled 300 unique prompts from the ShareGPT dataset (excluding those used in training) and conducted a comparative evaluation between the teacher and student models using GPT-4 as a judge (i.e., LLM-as-Judge). The model win rate was then calculated \cite{aligner} as:
    $$
    \omega=\frac{N_w-N_l}{N_w+N_l+N_e}\times 100\%
    $$
    where $N_w, N_l,N_e$ are the win/lose/tie count of the student vs the teacher.  
\paragraph{MT-Bench} \cite{vicuna} is a multi-turn benchmark consisting of 160 questions spanning eight diverse knowledge domains. A model under evaluation needs to respond to an initial query, followed by a second reply to a predefined follow-up question. GPT-4\footnote{All mentions of GPT-4 in this paper refer to the OpenAI GPT-4o-08-06 model.} evaluates each response on a scale of 1 to 10, with the final score being the average of the two turns.

\paragraph{AlpacaEval} \cite{alpaca_eval} is a single-turn benchmark in which the model generates responses to 805 questions covering various topics, with a focus on helpfulness. Following  \cite{zephyr, alpaca_eval}, the evaluation metrics for AlpacaEval are the mean and standard deviation of the tested model's win rate against a baseline model, text-davinci-003, as judged by GPT-4.

\subsection{Experimental Settings}
\paragraph{Teacher/Student Models} We conduct experiments in two settings: (1) capability recovery for pruned models with the help of the full-size LLM; and (2) enhancing the conversational ability of smaller models by distillation from larger LLM within an LLM Series.  

For pruned models, we use Vicuna1.5-7B \cite{vicuna}, an open-source conversational model fine-tuned from LLaMA2 \cite{llama2} on user-contributed conversations from ShareGPT, as our teacher model. The student model is obtained by applying LLM-Pruner \cite{llm-pruner}, a structured pruning method that identifies groups of hidden states that activate together during the forward pass. We prune 50\% and 20\% of the parameters, yielding the models Vicuna-7B\textsubscript{0.5} and Vicuna-7B\textsubscript{0.8}, respectively.  

For the LLM Series setting, we use the Qwen2.5 Series \cite{qwen2.5} for experiments. Specifically, Qwen2.5-7B-Instruct serves as the teacher LLM, while the 1.5B and 0.5B versions are used as student LLMs for distillation.

\paragraph{Baselines} To assess the impact of the teacher's distribution on the distillation process, we compare our proposed methods, \textit{rDPO} and \textit{daDPO}, against dSFT and dDPO, which do not leverage the teacher's distribution. Additionally, to evaluate the effectiveness of our approach, we compare \textit{daDPO} with alternative methods for incorporating the teacher's distribution, including dSFT+KL and dDPO+KL. More details on the baseline methods are provided in Section \ref{sec:baseline} and Appendix \ref{sec:appendix_baseline}.

% We consider several baselines as follows:
% \begin{enumerate}
%     \item dSFT \cite{zephyr}: We use \(\{(x, y_w)\}\) for Supervised Fine-Tuning of the student models.
%     \item dSFT+KL \cite{prune_kd,minillm_kl}: We use \(\{(x, y_w)\}\) to supervise and fine-tune the student model, and incorporate a KL divergence loss at the token level of $y_w$  between the teacher model and student model into SFT loss. 
%     \item dDPO \cite{zephyr,qdpo,smallbutfunny}: We use  \(\{x, y_w, y_l\}\) to optimize the student model after dSFT by DPO algorithm. 
%     \item dDPO+KL: We use  \(\{x, y_w, y_l\}\) to optimize the student model after dSFT, and then add a KL divergence loss at the token level of the teacher model response $y_w$ between two models into DPO loss. 
% \end{enumerate}

\paragraph{Implementation Details}
Like dDPO, \textit{daDPO} is applied after dSFT. We implement the \textit{daDPO} algorithm using the OpenRLHF repository\footnote{https://github.com/OpenRLHF/OpenRLHF}, requiring only a few additional lines of code. We tune \(\beta_1\) in \textit{daDPO} and \(\beta\) in all DPO baselines within the range \([0.01, 0.1, 1.0]\), while \(\beta_2\) in \textit{daDPO} is tuned from \([0.001, 0.01, 0.1, 1.0]\). Based on preliminary experiments, we set the batch size to 32 and the learning rate to \(5 \times 10^{-7}\) for all experiments. 

\subsection{Experimental Results}
\subsubsection{How does the teacher's distribution  
help the student model?}
\begin{table}[t]
\resizebox{\columnwidth}{!}{%
\begin{tabular}{@{}l|c|c|c@{}}
\toprule
\textbf{Method}                            & \textbf{MT-Bench} & \textbf{AlpacaEval} & \textbf{Indomain} \\ \midrule
Vicuna-7B                                   & 4.74                     & 65.71\%\textsubscript{1.67}                   & 0.0\%                   \\
\midrule
Vicuna-7B\textsubscript{0.8}                            & 3.06                     & 26.83\%\textsubscript{1.56}     & -50.6\%                    \\
\quad +dSFT                             & 3.68                     & 43.97\%\textsubscript{1.75}                      & -40.0\%                      \\
\quad +dDPO                             & 3.78                     & 50.94\%\textsubscript{1.77}                     & -31.0\%                      \\
\rowcolor{light-gray}\quad+rDPO & 3.82                     & 53.73\%\textsubscript{1.77}                      & -9.0\%                     \\
\rowcolor{light-gray}\quad+daDPO                             &    3.84                      &54.04\%\textsubscript{1.76}                            &-7.3\%                           \\
 \midrule
Vicuna-7B\textsubscript{0.5}                            & 1.31                     & 0.49\%\textsubscript{0.23}     & -86.6\%                    \\
\quad +dSFT                             & 2.00                    & 8.71\%\textsubscript{0.99}                      & -81.6\%                      \\
\quad +dDPO                             & 2.24                     & 12.32\%\textsubscript{1.16}                     & -73.6\%                 \\
\rowcolor{light-gray}\quad+rDPO & 2.24                     & 11.28\%\textsubscript{1.12}                      & -77.0\%                      \\
\rowcolor{light-gray}\quad+daDPO                             & 2.33                         &12.89\%\textsubscript{1.16}                            &-68.0\%                           \\
\bottomrule
\end{tabular}
}
\caption{Performance of daDPO on pruned Vicuna-7B compared to different baselines without distribution. Here, Vicuna-7B\textsubscript{0.8} and Vicuna-7B\textsubscript{0.5} refer to the model with 20\% and 50\% of the parameters pruned respectively.}

\label{tab:pruned-result}
\end{table}

\begin{table}[t]
\resizebox{\columnwidth}{!}{%
\begin{tabular}{@{}l|c|c|c@{}}
\toprule
\textbf{Method}                            & \textbf{MT-Bench} & \textbf{AlpacaEval} & \textbf{Indomain} \\ \midrule
Qwen2.5-7B                                   & 7.42                     & 93.08\%\textsubscript{1.67}                   & 0.0\%                   \\
\midrule
Qwen2.5-1.5B                          & 5.28                     & 65.50\%\textsubscript{1.56}     &-61.7\%                     \\
\quad +dSFT                             & 5.36                     & 74.35\%\textsubscript{1.75}                      &-7.3\%                       \\
\quad +dDPO                             & 5.97                     &78.85\%\textsubscript{1.44}                      &0.7\%                      \\
\rowcolor{light-gray}\quad+rDPO &5.74 &78.94\%\textsubscript{1.52}  &  -5.0\%  \\
\rowcolor{light-gray}\quad+daDPO                             &  6.10                        &81.49\%\textsubscript{1.36}                           &14.0\%                           \\
 \midrule
Qwen2.5-0.5B                            & 3.65                     & 37.82\%\textsubscript{1.71}     & -75.0\%                    \\
\quad +dSFT                             & 3.71                    & 41.15\%\textsubscript{1.73}                      & -66.3\%                      \\
\quad +dDPO                             & 3.72                     & 42.27\%\textsubscript{1.74}                     & -63.3\%                      \\
\rowcolor{light-gray}\quad+rDPO&3.70 & 43.41\%\textsubscript{1.74}  & -48.7\% \\
\rowcolor{light-gray}\quad+daDPO                             &  3.80                        &45.09\%\textsubscript{1.75}                            &-55.3\%                           \\
\bottomrule
\end{tabular} 
}
\caption{Performance of daDPO in enhancing smaller LLMs within an LLM Series. In this case, the teacher model is Qwen2.5-7B, while the student models are Qwen2.5-1.5B and Qwen2.5-0.5B, respectively.}
\label{tab:qwen-series-results}
\end{table}

\paragraph{Performance of \textit{daDPO}} Tables \ref{tab:pruned-result} and \ref{tab:qwen-series-results} demonstrate that \textit{daDPO} consistently outperforms other baselines in both experimental settings. For Vicuna-7B\textsubscript{0.8} (20\% pruned) in Table \ref{tab:pruned-result}, \textit{daDPO} shows significant improvements over dDPO on MT-Bench (3.84 vs. baseline dDPO's 3.78), AlpacaEval (54.04\% vs. 50.94\%), and in-domain evaluation (-7.3\% vs. -31.0\%). For the more challenging Vicuna-7B\textsubscript{0.5} (50\% pruned), \textit{daDPO} maintains its advantage, achieving the best results across all three metrics. In Table \ref{tab:qwen-series-results}, for Qwen2.5-1.5B, \textit{daDPO} achieves the best performance across datasets, with a notable in-domain win rate of 14.0\% (versus dDPO's 0.7\%). This positive win rate is especially significant, as it shows that, without additional labeling cost, the daDPO-enhanced 1.5B model can occasionally outperform its 7B teacher model. This could be due to the rank information itself providing valuable signals that the teacher model had not encountered before. Similar improvements are observed for the smaller Qwen2.5-0.5B model, where \textit{daDPO} achieves an AlpacaEval score of 45.09\% (vs. dDPO's 42.27\%) and an in-domain win rate of -53.3\% (compared to dDPO's -63.3\%).

\begin{figure}[t]
\centering
  \includegraphics[width=0.98\linewidth]{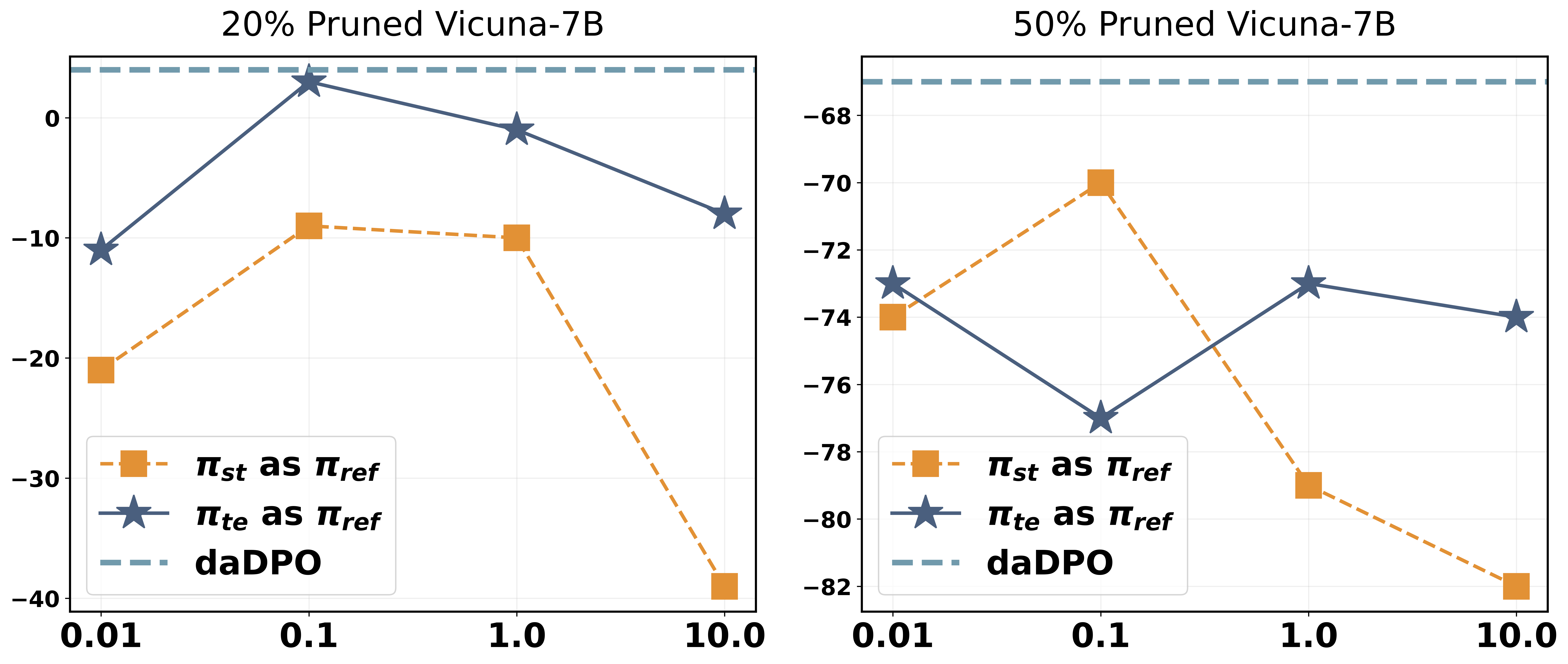}
  \caption{Comparison of in-domain performance of 100 test samples between DPO with different models (teacher v.s. student) as reference policy. Here $x$ is different $\beta$ values and $y$ is in-domain score}
  \label{fig:in-domain}
\end{figure}

\paragraph{Performance of \textit{rDPO}} While integrating the teacher's distribution proves beneficial, the role of the student's distribution as the reference model is also important. As indicated in Table~\ref{tab:pruned-result}, rDPO outperforms dDPO at a 20\% pruning rate but suffers from degraded performance at 50\% pruning. To further investigate this, we analyzed the in-domain performance on 100 test samples for both dDPO and rDPO across varying values of $\beta$. The results in Figure~\ref{fig:in-domain} show that while replacing the reference policy (i.e., using $\pi_{te}$ as $\pi_{ref}$) can improve performance in some cases, discarding the reference policy constraint leads to instability, especially as the model difference increases (as in the heavily pruned case). This suggests that as the difference between the two models increases, the importance of the student distribution as the reference policy for ensuring stable training becomes more pronounced. %This empirical evidence reinforces our hypothesis that the reference policy in daDPO plays a crucial role that should not be overlooked.

\subsubsection{What is the effective way to integrate the teacher's distribution?}

\begin{figure}[t]
\centering
\begin{subfigure}{0.99\linewidth} % 控制子图的宽度
        \centering
        \includegraphics[width=\textwidth]{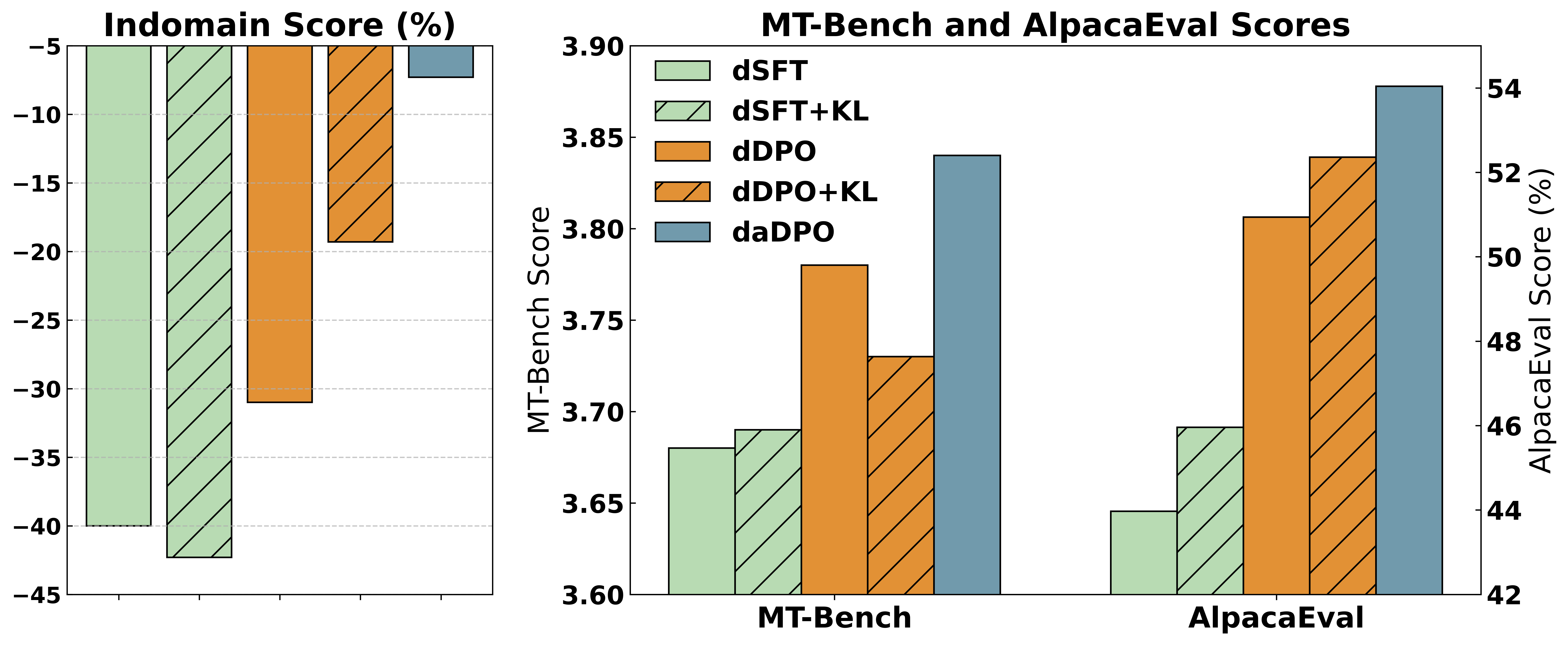} % 替换为你的图片
        \caption{20\% pruned Vicuna-7B}
        \label{fig:subfig1}
    \end{subfigure}
    \\
    % 子图2
    \begin{subfigure}{0.99\linewidth}
        \centering
        \includegraphics[width=\textwidth]{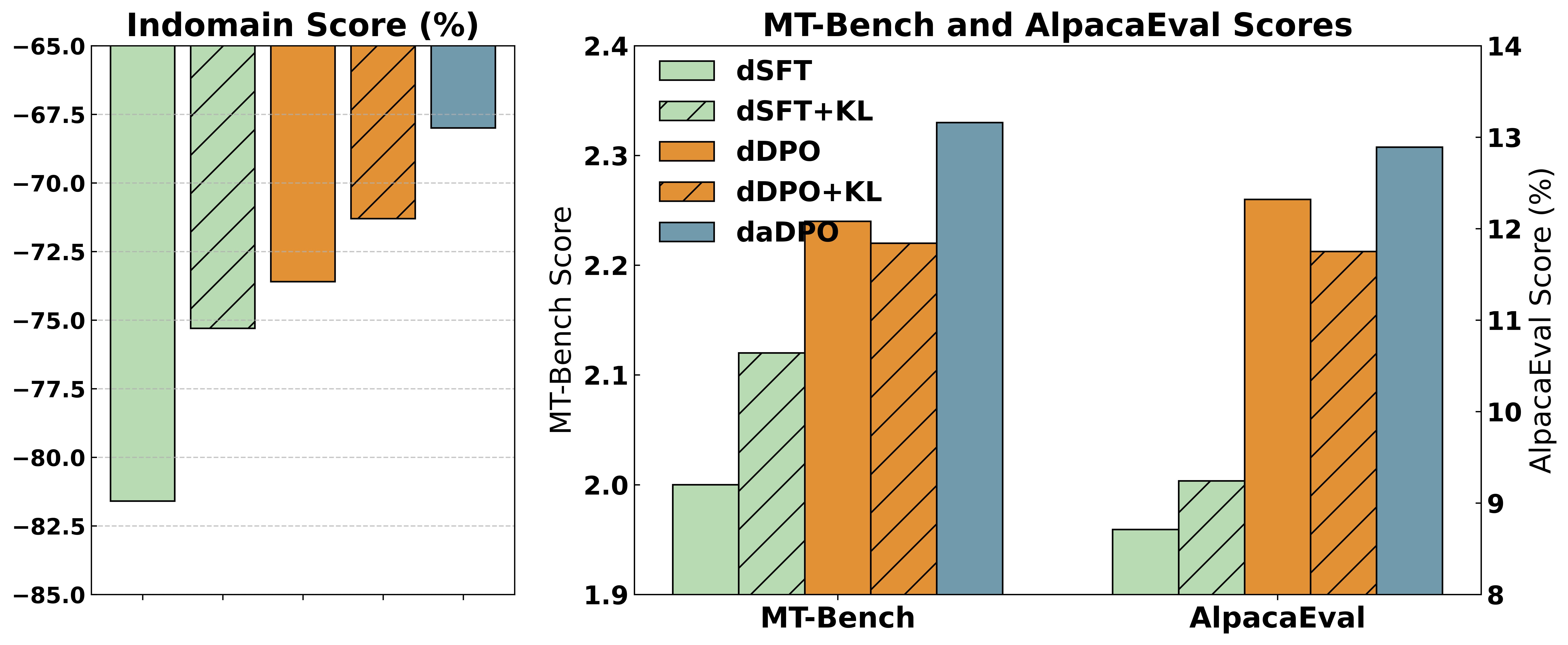} % 替换为你的图片
        \caption{50\% pruned Vicuna-7B}
        \label{fig:subfig2}
    \end{subfigure}
  \caption{Baselines with distribution information on Vicuna7B-20\%pruned and Vicuna7B-50\%pruned. It shows that \textit{daDPO} consistently outperforms directly adding distillation KL loss into SFT or DPO loss.}
  \label{fig:vicuna-dist}
\end{figure}

As discussed in Section \ref{sec:dadpo}, a straightforward approach to integrating the teacher's distribution is by introducing KL loss (Eq. \ref{kl-loss}) into dSFT and dDPO. We investigate whether our approach, \textit{daDPO}, offers a more effective way to incorporate the teacher's distribution for recovering the capability of pruned models. Figure \ref{fig:vicuna-dist} illustrates the performance gap between the pruned Vicuna-7B models using dDPO, dDPO+KL, dSFT, and dSFT+KL. The results show that adding KL loss to SFT and DPO can improve performance in some cases, but not consistently. For example, with the 20\% pruned Vicuna-7B, dSFT+KL results in a lower in-domain score than dSFT alone, and dDPO+KL yields a worse MT-Bench score than dDPO. In contrast, our proposed \textit{daDPO} method, with its integrated approach, achieves the best scores across all metrics.

%\subsection{Further Analysis}
\subsubsection{Does enhancing chat capability make the student forget other abilities?}
Alignment tax refers to the phenomenon where enhancing conversational ability can lead to degraded performance on other abilities. This issue has been widely observed with RLHF \cite{mitigating_tax} and with distillation for quantized models \cite{qdpo}. In this study, we examine whether our proposed method exhibits a higher alignment tax compared to dSFT and dDPO when applied to the Qwen2.5 Series of LLMs.  

The results in Table \ref{tab:mmlu} show that improving conversational abilities through \textit{daDPO} does not come at much expense of academic task performance. Specifically, compared to dDPO, \textit{daDPO} preserves the performance of Qwen2.5-1.5B on MMLU and GSM8K while slightly improving the scores on ARC\_C (56.1\% vs. 55.5\% for dDPO) and Hellaswag (68.3\% vs. 68.1\% for dDPO). Similarly, \textit{daDPO} enhances Qwen2.5-0.5B on three out of four tasks, including MMLU (46.8\% vs. 46.4\% for dDPO), ARC\_C (35.9\% vs. 35.4\% for dDPO), and Hellaswag (51.1\% vs. 50.7\% for dDPO). On GSM8K, alignment tax is observed with both dDPO and \textit{daDPO}. However, \textit{daDPO} does not introduce any additional penalty, maintaining performance close to that of dDPO.

\begin{table}[t]
\resizebox{\columnwidth}{!}{%
\begin{tabular}{@{}l|c|c|c|c@{}}
\toprule
\textbf{Method}                            & \textbf{MMLU} & \textbf{ARC\_C} & \textbf{GSM8K} & \textbf{Hellaswag} \\ \midrule
Qwen2.5-7B                                   &            74.2         &     66.9                &        82.4          &  81.4 \\
\midrule
Qwen2.5-1.5B                          &60.2                      &  54.8    &    53.1&67.0            \\
\quad +dSFT                             &  60.3                    &      54.1                &  53.1 & 67.0                    \\
\quad +dDPO                             &    60.4                  &   55.6                   & 52.4 & 68.1                  \\
\quad+daDPO                             &              60.4            &  56.1                          &   52.4 &68.3                    \\
 \midrule
Qwen2.5-0.5B                            &          46.7            &   36.1     &   31.8 & 51.0                  \\
\quad +dSFT                             &        46.6             &    35.8                  &    34.2    &    50.9               \\
\quad +dDPO                             &          46.4            &    35.4                &            30.6 & 50.7          \\
\quad+daDPO                             &         46.8                 &      35.9                      &         30.5 & 51.1                \\
\bottomrule
\end{tabular}%
}
\caption{The alignment tax is maintained at a low rate.}
\label{tab:mmlu}
\end{table}

%On MMLU, the performance remains stable at 60.4\% for both model sizes. For ARC\_C, daDPO achieves the best results among all methods (56.1\% for both sizes), showing a small but consistent improvement over dDPO (55.6\% and 55.3\%). On GSM8k, daDPO matches dDPO's performance for Qwen2.5-1.5b (52.4\%) and slightly improves it for Qwen2.5-0.5b (45.8\% vs 45.4\%). Similarly, for Hellaswag, daDPO shows marginal improvements (68.3\%) compared to dDPO (68.1\%) for both model sizes. These results suggest that daDPO's approach to enhancing conversational abilities through distribution-aware preference optimization can be achieved while preserving or slightly strengthening the model's capabilities on structured reasoning and knowledge-intensive tasks. 

\subsubsection{How does the teacher's size matter?}
\begin{figure}[t]
\centering
  \includegraphics[width=0.88\linewidth]{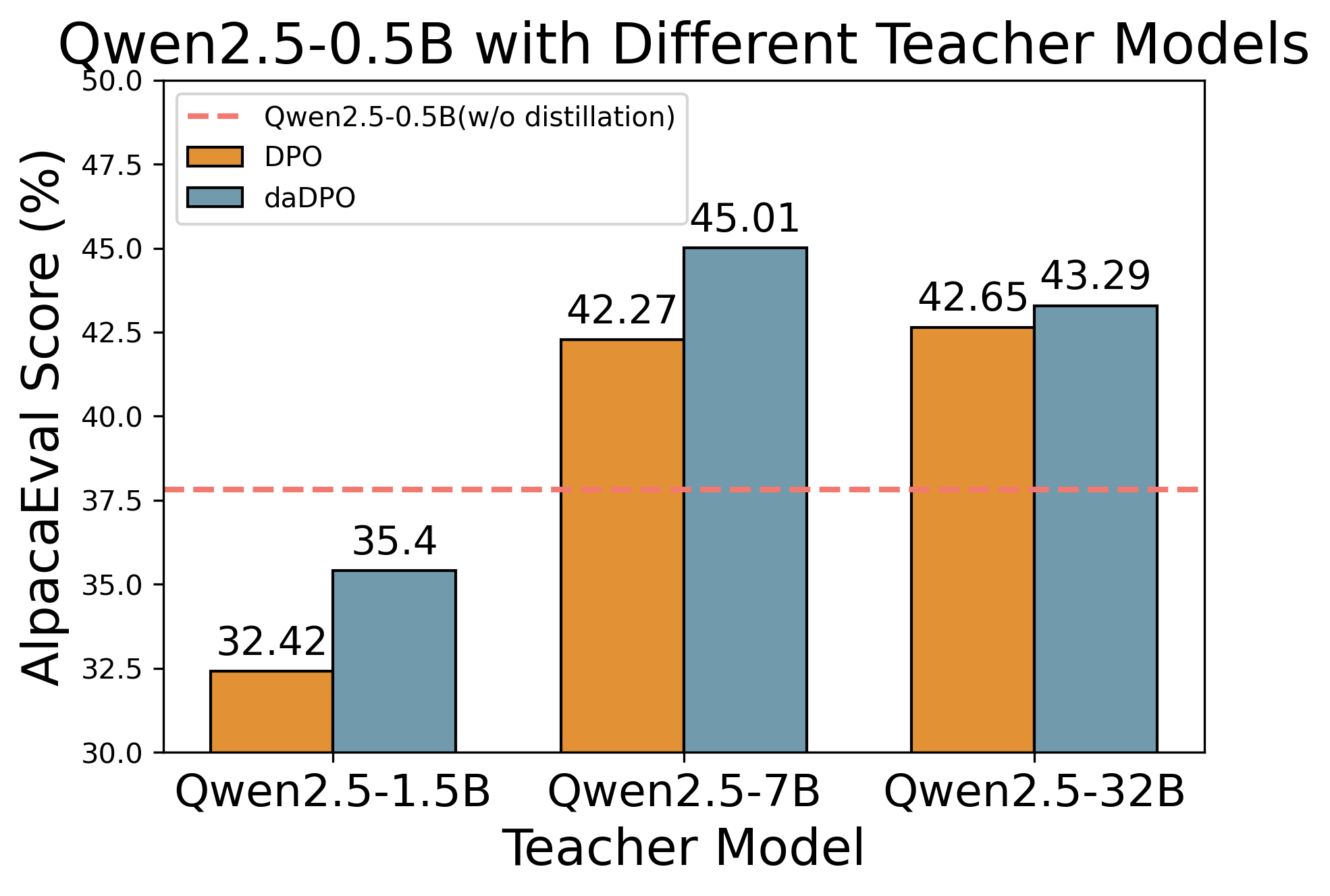}  
  \caption{
  Comparison of AlpacaEval scores for Qwen2.5-0.5B student model using different teacher models (Qwen2.5-1.5B, Qwen2.5-7B, Qwen2.5-32B) with dDPO and \textit{daDPO} methods. Results show that \textit{daDPO} outperforms dDPO, with the highest win rate achieved using Qwen2.5-7B as the teacher model.
  }
  \label{fig:different-teacher}
\end{figure}

We conducted experiments using Qwen2.5-0.5B as the student model and the 1.5B, 7B, and 32B versions as teacher models. The results in Figure~\ref{fig:different-teacher} show that the 7B version is a more suitable teacher for the 0.5B one in comparison with the 1.5B and the 32B. This highlights the importance of balancing the teacher's capability with the teacher-student knowledge gap. More specifically, since the 32B and 7B teachers achieve similar AlpacaEval scores (95.46 vs. 93.08), the smaller 7B teacher—exhibiting less distribution shift relative to the 1.5B student—facilitates more effective distillation. In contrast, the 1.5B teacher's advantage over the 0.5B student (65.50 vs. 37.82) is less significant, indicating a higher noise rate in the preference dataset that may hinder distillation performance. Nevertheless, \textit{daDPO} consistently maintains its advantage over dDPO.

% \torevise{
% \subsubsection{How does it matter if the teacher and student are from different families?}
% We conducted experiments using LLama3.2-1B as the student model and Qwen2.5-7b as teacher models. The results in Table~\ref{fig:different-series-results} show that the student's performance is significantly influenced by the teacher model's quality and compatibility. Despite being from different families, the student can still achieve positive results if the teacher provides effective guidance. This suggests that cross-family model pairing is viable with appropriate teacher selection.
% }

\begin{table}[t]
\resizebox{\columnwidth}{!}{%
\begin{tabular}{@{}l|c|c|c@{}}
\toprule
\textbf{Method}                            & \textbf{MT-Bench} & \textbf{AlpacaEval} & \textbf{Indomain} \\ \midrule
Qwen2.5-7B                                   & 7.42                     & 93.08\%\textsubscript{1.67}                   & 0.0\%                   \\
\midrule
LLama3.2-1B                          & 4.37                     & 17.64\%\textsubscript{1.34}     &-72.3\%                     \\
\quad +dSFT                             & 4.68                     & 63.10\%\textsubscript{1.70}                      &-38.0\%                       \\
\quad +dDPO                             & 4.71                     &65.34\%\textsubscript{1.67}                      &-36.3\%                      \\
\rowcolor{light-gray}\quad+rDPO                             & 4.69                        &64.10\%\textsubscript{1.69}                           &-39.3\%                           \\
\rowcolor{light-gray}\quad+daDPO                             &  4.99                        &65.71\%\textsubscript{1.67}                           &-36.0\%                           \\
 
\bottomrule
\end{tabular} 
}
\caption{Performance of daDPO when the teacher is Qwen2.5-7B, and the student model is LLama3.2-1B.}
\label{tab:different-series-results}
\end{table}

\subsubsection{Ablation study on $\beta_1$ and $\beta_2$}
\begin{figure}[t]
\centering
  \includegraphics[width=0.98\linewidth]{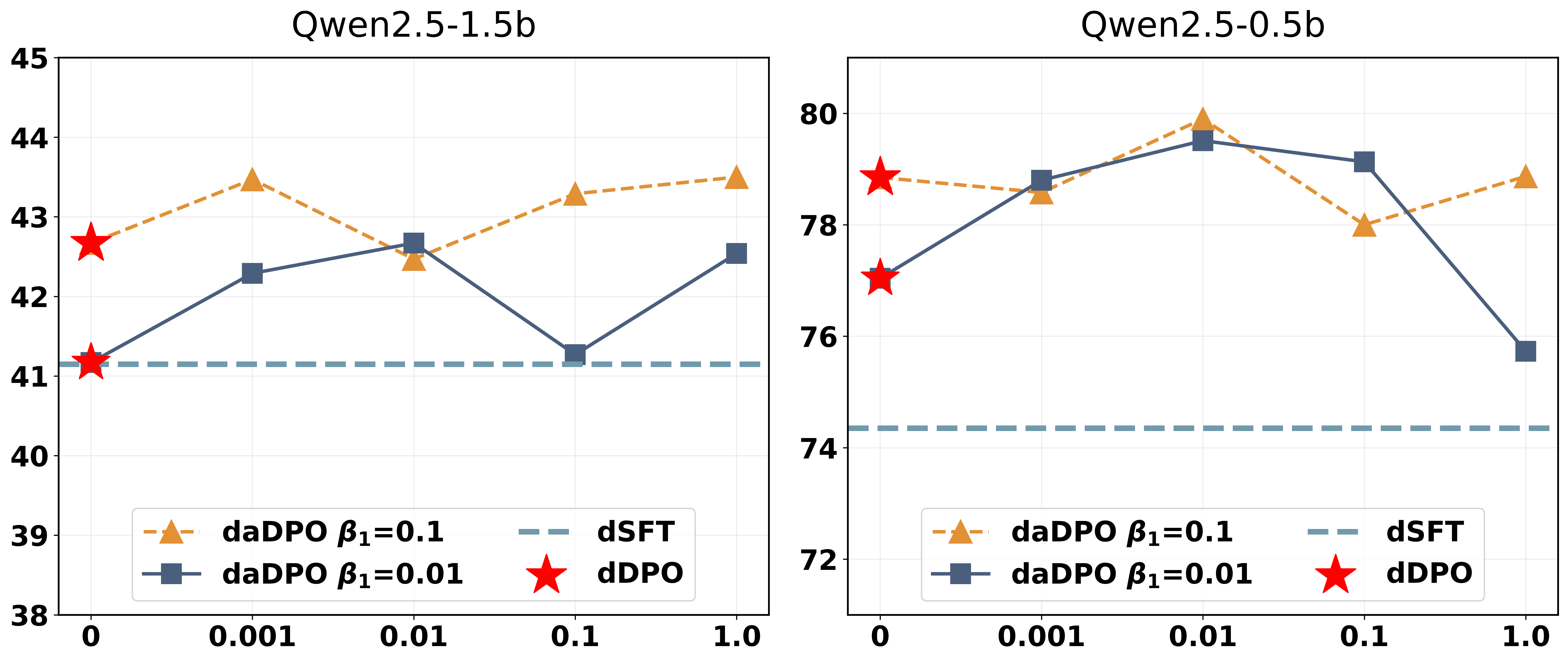}
  \caption{Ablation study on the impact of \(\beta_1\) (for reference model KL) and \(\beta_2\)(for teacher model KL) in \textit{daDPO}. Here y-axis shows the AlpacaEval scores and x-axis shows the value of $\beta$.  
  % It shows that: (1). Both dDPO and daDPO outperform dSFT across different values of $\beta_1$ and $\beta_2$. (2). Gradually increasing $\beta_2$ up to 1 consistently leads to better performance compared to dDPO (i.e., $\beta_2$= 0) but with some peak points in the middle.
  }
  \label{fig:beta_qwen}
\end{figure}

We adjust the values of $\beta_1$ and $\beta_2$ on the Qwen2.5 Series to observe how they affect the model's performance. Figure \ref{fig:beta_qwen} provides a visual comparison of the AlpacaEval scores for the 0.5B and 1.5B versions of Qwen2.5.

% \begin{figure}[t]
% \centering
%   \includegraphics[width=0.48\linewidth]{latex/qwen0.5b_beta1_beta2_bybeta2.png}
%   \includegraphics[width=0.48\linewidth]{latex/qwen1.5b_beta1_beta2_bybeta2.png}
%   \caption{Ablation study on the impact of $\beta_1$ and $\beta_2$ in daDPO. The top graph shows the AlpacaEval scores for Qwen2.5-0.5b at different ratios of $\beta_2/\beta_1$, with $\beta_1$ fixed at 0.01 and 0.1. The bottom graph illustrates the same for Qwen2.5-1.5b. The results indicate that the model's performance peaks at an optimal $\beta_2/\beta_1$ ratio (around 1.0 for $\beta_1 = 0.01$), beyond which further increases lead to a decline in performance. This highlights the importance of balancing the reference policy constraint ($\beta_1$) and the teacher distribution information ($\beta_2$) to maximize conversational abilities.}
%   \label{fig:beta_qwen}
%   \legend{%
%     \includegraphics[width=1cm]{latex/qwen0.5b_beta1_beta2_bybeta2.png} DDPO $\beta_1=0.01$,%
%     \includegraphics[width=1cm]{latex/qwen1.5b_beta1_beta2_bybeta2.png} DPO $\beta_1=0.1$,%
%     ncols=2 % 分两列显示图例
%   }
% \end{figure}

The results from Figure \ref{fig:beta_qwen} suggests several insights:
(1) Both dDPO and \textit{daDPO} outperform dSFT across different values of $\beta_1$ and $\beta_2$, demonstrating the effectiveness of preference learning in distillation tasks; % The improvements are more pronounced on Qwen2.5-1.5B, likely because the distribution gap between it and the 7B teacher is smaller, making preference learning more stable. \cite{dpo} 
(2) In most cases, gradually increasing $\beta_2$ up to 1 consistently leads to better performance compared to dDPO (i.e., $\beta_2$= 0). However, the ratio between two KL terms should be carefully controlled and not be too large. When $\beta_2$ is excessively high, the reference model imposes weaker constraints, leading to less stable training and ultimately affecting performance.

\subsubsection{Dose daDPO work with models from different families?}
Our previous experiments focused on student-teacher pairs from the same model family.
To evaluate the broader applicability of our method, we conducted cross-family distillation experiments between Qwen2.5-7B and Llama3.2-1B. Results in Table \ref{tab:different-series-results} demonstrate that \textit{daDPO} outperforms dDPO and dSFT on all three benchmarks. The suboptimal performance of rDPO aligns with previous findings \cite{sjtu_policymodel}, which can be attributable to the increased model distance between student and teacher architectures across families. The results demonstrate that while rDPO becomes ineffective, \textit{daDPO maintains its functionality with models from different model families.}

\section{Conclusion}
In this work, we developed Distribution Aware DPO (\textit{daDPO}), a novel framework that integrates the rich distributional information of white-box teacher LLMs into the Direct Preference Optimization (DPO) distillation process. Our approach offers several advantages, including more stable ``white-box'' distillation, the potential for cross-LLM family distillation, and the ability to maintain DPO’s simplicity and theoretical guarantees. Our experiments on pruned models (Vicuna-7B) and smaller models in the Qwen2.5 Series demonstrated that \textit{daDPO} consistently outperforms existing baselines, achieving significant improvements in conversational performance without compromising downstream task capabilities (compared to dDPO). This work provides a flexible and theoretically grounded approach to knowledge distillation, enabling more efficient deployment of smaller, yet highly capable, language models in resource-constrained environments.

\section{Limitation}
Despite the effectiveness and novelty of the methodology in this paper, it has several limitations as follows:
\begin{enumerate}
    \item Even though we have made a step forward compared to the traditional distribution-based knowledge distillation method which requires the teacher model and student model to share the same tokenization vocabulary space, the proposed method still requires the teacher model to be white-box as we need the sentence possibility of it to calculate the loss.  
    \item The Scope of this paper is mostly limited to general chat ability distillation. In the experiment part, both training datasets and evaluations are limited to general chat ability, whether our method can be generalized in other specific areas like reasoning ability distillation is still unknown and needs further study.
\end{enumerate}
\bibliography{custom}

\begin{thebibliography}{42}
\providecommand{\natexlab}[1]{#1}

\bibitem[{Achiam et~al.(2023)Achiam, Adler, Agarwal, Ahmad, Akkaya, Aleman, Almeida, Altenschmidt, Altman, Anadkat et~al.}]{gpt4-2023}
Josh Achiam, Steven Adler, Sandhini Agarwal, Lama Ahmad, Ilge Akkaya, Florencia~Leoni Aleman, Diogo Almeida, Janko Altenschmidt, Sam Altman, Shyamal Anadkat, et~al. 2023.
\newblock Gpt-4 technical report.
\newblock \emph{arXiv preprint arXiv:2303.08774}.

\bibitem[{Agarwal et~al.(2024)Agarwal, Vieillard, Zhou, Stanczyk, Garea, Geist, and Bachem}]{onpolicy_distillation}
Rishabh Agarwal, Nino Vieillard, Yongchao Zhou, Piotr Stanczyk, Sabela~Ramos Garea, Matthieu Geist, and Olivier Bachem. 2024.
\newblock On-policy distillation of language models: Learning from self-generated mistakes.
\newblock In \emph{The Twelfth International Conference on Learning Representations}.

\bibitem[{Azar et~al.(2024)Azar, Guo, Piot, Munos, Rowland, Valko, and Calandriello}]{ipo}
Mohammad~Gheshlaghi Azar, Zhaohan~Daniel Guo, Bilal Piot, Remi Munos, Mark Rowland, Michal Valko, and Daniele Calandriello. 2024.
\newblock A general theoretical paradigm to understand learning from human preferences.
\newblock In \emph{International Conference on Artificial Intelligence and Statistics}, pages 4447--4455. PMLR.

\bibitem[{Claude(2024)}]{claude}
Claude. 2024.
\newblock Conversation with claude 3.5 sonnet.
\newblock Anthropic.
\newblock Anthropic AI assistant, version 3.5 Sonnet, knowledge cutoff April 2024.

\bibitem[{DeepSeek-AI et~al.(2025)DeepSeek-AI, Yang, Zhang, Song, Zhang, Xu, Zhu, Ma, Wang, Bi et~al.}]{deepseekr1}
Daya~Guo DeepSeek-AI, Dejian Yang, Haowei Zhang, Junxiao Song, Ruoyu Zhang, Runxin Xu, Qihao Zhu, Shirong Ma, Peiyi Wang, Xiao Bi, et~al. 2025.
\newblock Deepseek-r1: Incentivizing reasoning capability in llms via reinforcement learning.
\newblock \emph{arXiv preprint arXiv:2501.12948}.

\bibitem[{Ethayarajh et~al.(2024)Ethayarajh, Xu, Muennighoff, Jurafsky, and Kiela}]{ethayarajh2024kto}
Kawin Ethayarajh, Winnie Xu, Niklas Muennighoff, Dan Jurafsky, and Douwe Kiela. 2024.
\newblock Kto: Model alignment as prospect theoretic optimization.
\newblock \emph{arXiv preprint arXiv:2402.01306}.

\bibitem[{Gu et~al.()Gu, Yang, Ding, Zhao, and Tan}]{tanfei2_dch_scalinglaw}
Jiawei Gu, Zacc Yang, Chuanghao Ding, Rui Zhao, and Fei Tan.
\newblock {CMR} scaling law: Predicting critical mixture ratios for continual pre-training of language models.
\newblock In \emph{Proceedings of the 2024 Conference on Empirical Methods in Natural Language Processing}.

\bibitem[{Gu et~al.(2024)Gu, Dong, Wei, and Huang}]{minillm_kl}
Yuxian Gu, Li~Dong, Furu Wei, and Minlie Huang. 2024.
\newblock Minillm: Knowledge distillation of large language models.
\newblock In \emph{The Twelfth International Conference on Learning Representations}.

\bibitem[{Hinton et~al.(2015)Hinton, Vinyals, and Dean}]{hinton2015distilling}
Geoffrey Hinton, Oriol Vinyals, and Jeff Dean. 2015.
\newblock Distilling the knowledge in a neural network.
\newblock \emph{arXiv preprint arXiv:1503.02531}.

\bibitem[{Ji et~al.(2024)Ji, Chen, Lou, Hong, Zhang, Pan, Qiu, Dai, and Yang}]{aligner}
Jiaming Ji, Boyuan Chen, Hantao Lou, Donghai Hong, Borong Zhang, Xuehai Pan, Tianyi Qiu, Juntao Dai, and Yaodong Yang. 2024.
\newblock Aligner: Efficient alignment by learning to correct.
\newblock In \emph{The Thirty-eighth Annual Conference on Neural Information Processing Systems}.

\bibitem[{Lee et~al.(2024)Lee, Park, Hong, Kim, Chang, and Choi}]{qdpo}
Janghwan Lee, Seongmin Park, Sukjin Hong, Minsoo Kim, Du-Seong Chang, and Jungwook Choi. 2024.
\newblock Improving conversational abilities of quantized large language models via direct preference alignment.
\newblock In \emph{Proceedings of the 62nd Annual Meeting of the Association for Computational Linguistics (Volume 1: Long Papers)}, pages 11346--11364.

\bibitem[{Li et~al.(2023{\natexlab{a}})Li, Zhao, and Tan}]{tanfei4_segmentation}
Dedong Li, Rui Zhao, and Fei Tan. 2023{\natexlab{a}}.
\newblock Cwseg: An efficient and general approach to chinese word segmentation.
\newblock In \emph{Proceedings of the 61st Annual Meeting of the Association for Computational Linguistics (Volume 5: Industry Track)}, pages 1--10.

\bibitem[{Li et~al.(2023{\natexlab{b}})Li, Zhang, Dubois, Taori, Gulrajani, Guestrin, Liang, and Hashimoto}]{alpaca_eval}
Xuechen Li, Tianyi Zhang, Yann Dubois, Rohan Taori, Ishaan Gulrajani, Carlos Guestrin, Percy Liang, and Tatsunori~B. Hashimoto. 2023{\natexlab{b}}.
\newblock Alpacaeval: An automatic evaluator of instruction-following models.
\newblock \url{https://github.com/tatsu-lab/alpaca_eval}.

\bibitem[{Li et~al.(2024)Li, Gu, Dong, Wang, Cheng, and Wei}]{li2024direct}
Yixing Li, Yuxian Gu, Li~Dong, Dequan Wang, Yu~Cheng, and Furu Wei. 2024.
\newblock Direct preference knowledge distillation for large language models.
\newblock \emph{arXiv preprint arXiv:2406.19774}.

\bibitem[{Lin et~al.(2024)Lin, Lin, Xiong, Diao, Liu, Zhang, Pan, Wang, Hu, Zhang et~al.}]{mitigating_tax}
Yong Lin, Hangyu Lin, Wei Xiong, Shizhe Diao, Jianmeng Liu, Jipeng Zhang, Rui Pan, Haoxiang Wang, Wenbin Hu, Hanning Zhang, et~al. 2024.
\newblock Mitigating the alignment tax of rlhf.
\newblock In \emph{Proceedings of the 2024 Conference on Empirical Methods in Natural Language Processing}, pages 580--606.

\bibitem[{Liu et~al.(2024{\natexlab{a}})Liu, Feng, Wang, Wang, Liu, Zhao, Dengr, Ruan, Dai, Guo et~al.}]{liu2024deepseekv2}
Aixin Liu, Bei Feng, Bin Wang, Bingxuan Wang, Bo~Liu, Chenggang Zhao, Chengqi Dengr, Chong Ruan, Damai Dai, Daya Guo, et~al. 2024{\natexlab{a}}.
\newblock Deepseek-v2: A strong, economical, and efficient mixture-of-experts language model.
\newblock \emph{arXiv preprint arXiv:2405.04434}.

\bibitem[{Liu et~al.(2024{\natexlab{b}})Liu, Liu, and Cohan}]{sjtu_policymodel}
Yixin Liu, Pengfei Liu, and Arman Cohan. 2024{\natexlab{b}}.
\newblock Understanding reference policies in direct preference optimization.
\newblock \emph{arXiv preprint arXiv:2407.13709}.

\bibitem[{Lu et~al.(2022)Lu, Zhu, Han, Zhao, Mac~Namee, and Tan}]{tanfei3_pretain}
Jinghui Lu, Dongsheng Zhu, Weidong Han, Rui Zhao, Brian Mac~Namee, and Fei Tan. 2022.
\newblock What makes pre-trained language models better zero-shot learners?
\newblock \emph{Association for Computational Linguistics ACL 2023}.

\bibitem[{Ma et~al.(2023)Ma, Fang, and Wang}]{llm-pruner}
Xinyin Ma, Gongfan Fang, and Xinchao Wang. 2023.
\newblock Llm-pruner: On the structural pruning of large language models.
\newblock \emph{Advances in neural information processing systems}, 36:21702--21720.

\bibitem[{Meng et~al.(2024)Meng, Xia, and Chen}]{meng2024simpo}
Yu~Meng, Mengzhou Xia, and Danqi Chen. 2024.
\newblock Simpo: Simple preference optimization with a reference-free reward.
\newblock \emph{Advances in Neural Information Processing Systems}.

\bibitem[{Muralidharan et~al.(2024)Muralidharan, Sreenivas, Joshi, Chochowski, Patwary, Shoeybi, Catanzaro, Kautz, and Molchanov}]{prune_kd}
Saurav Muralidharan, Sharath~Turuvekere Sreenivas, Raviraj~Bhuminand Joshi, Marcin Chochowski, Mostofa Patwary, Mohammad Shoeybi, Bryan Catanzaro, Jan Kautz, and Pavlo Molchanov. 2024.
\newblock Compact language models via pruning and knowledge distillation.
\newblock In \emph{The Thirty-eighth Annual Conference on Neural Information Processing Systems}.

\bibitem[{Ouyang et~al.(2022)Ouyang, Wu, Jiang, Almeida, Wainwright, Mishkin, Zhang, Agarwal, Slama, Ray et~al.}]{instructgpt2022}
Long Ouyang, Jeffrey Wu, Xu~Jiang, Diogo Almeida, Carroll Wainwright, Pamela Mishkin, Chong Zhang, Sandhini Agarwal, Katarina Slama, Alex Ray, et~al. 2022.
\newblock Training language models to follow instructions with human feedback.
\newblock \emph{Advances in Neural Information Processing Systems}.

\bibitem[{Rafailov et~al.(2024)Rafailov, Hejna, Park, and Finn}]{qfunction}
Rafael Rafailov, Joey Hejna, Ryan Park, and Chelsea Finn. 2024.
\newblock From r to q: Your language model is secretly a q-function.
\newblock \emph{First Conference on Language Modeling}.

\bibitem[{Rafailov et~al.(2023)Rafailov, Sharma, Mitchell, Manning, Ermon, and Finn}]{dpo}
Rafael Rafailov, Archit Sharma, Eric Mitchell, Christopher~D Manning, Stefano Ermon, and Chelsea Finn. 2023.
\newblock Direct preference optimization: Your language model is secretly a reward model.
\newblock \emph{Advances in Neural Information Processing Systems}, 36.

\bibitem[{Ravi et~al.(2024)Ravi, Huber, Shrivastava, Sagar, Aly, Shwartz, and Einolghozati}]{smallbutfunny}
Sahithya Ravi, Patrick Huber, Akshat Shrivastava, Aditya Sagar, Ahmed Aly, Vered Shwartz, and Arash Einolghozati. 2024.
\newblock Small but funny: A feedback-driven approach to humor distillation.
\newblock \emph{Proceedings of the 62nd Annual Meeting of the Association for Computational Linguistics}.

\bibitem[{Sanh(2019)}]{sanh2019distilbert}
V~Sanh. 2019.
\newblock Distilbert, a distilled version of bert: smaller, faster, cheaper and lighter.
\newblock In \emph{Proceedings of Thirty-third Conference on Neural Information Processing Systems (NIPS2019)}.

\bibitem[{Schulman et~al.(2017)Schulman, Wolski, Dhariwal, Radford, and Klimov}]{ppo}
John Schulman, Filip Wolski, Prafulla Dhariwal, Alec Radford, and Oleg Klimov. 2017.
\newblock Proximal policy optimization algorithms.
\newblock \emph{arXiv preprint arXiv:1707.06347}.

\bibitem[{Stiennon et~al.(2020)Stiennon, Ouyang, Wu, Ziegler, Lowe, Voss, Radford, Amodei, and Christiano}]{summarize2020}
Nisan Stiennon, Long Ouyang, Jeffrey Wu, Daniel Ziegler, Ryan Lowe, Chelsea Voss, Alec Radford, Dario Amodei, and Paul~F Christiano. 2020.
\newblock Learning to summarize with human feedback.
\newblock \emph{Advances in Neural Information Processing Systems}.

\bibitem[{Sun et~al.(2025{\natexlab{a}})Sun, Sun, Peng, and Ye}]{sun2025mitigating}
Hai-Long Sun, Zhun Sun, Houwen Peng, and Han-Jia Ye. 2025{\natexlab{a}}.
\newblock Mitigating visual forgetting via take-along visual conditioning for multi-modal long cot reasoning.
\newblock In \emph{ACL}.

\bibitem[{Sun et~al.(2025{\natexlab{b}})Sun, Zhou, Li, Lu, Yi, Chen, Xu, Luo, Zhang, Zhan et~al.}]{sun2025parrot}
Hai-Long Sun, Da-Wei Zhou, Yang Li, Shiyin Lu, Chao Yi, Qing-Guo Chen, Zhao Xu, Weihua Luo, Kaifu Zhang, De-Chuan Zhan, et~al. 2025{\natexlab{b}}.
\newblock Parrot: Multilingual visual instruction tuning.
\newblock In \emph{ICML}.

\bibitem[{Taori et~al.(2023)Taori, Gulrajani, Zhang, Dubois, Li, Guestrin, Liang, and Hashimoto}]{alpaca}
Rohan Taori, Ishaan Gulrajani, Tianyi Zhang, Yann Dubois, Xuechen Li, Carlos Guestrin, Percy Liang, and Tatsunori~B Hashimoto. 2023.
\newblock Stanford alpaca: An instruction-following llama model.

\bibitem[{Touvron et~al.(2023)Touvron, Martin, Stone, Albert, Almahairi, Babaei, Bashlykov, Batra, Bhargava, Bhosale et~al.}]{llama2}
Hugo Touvron, Louis Martin, Kevin Stone, Peter Albert, Amjad Almahairi, Yasmine Babaei, Nikolay Bashlykov, Soumya Batra, Prajjwal Bhargava, Shruti Bhosale, et~al. 2023.
\newblock Llama 2: Open foundation and fine-tuned chat models.
\newblock \emph{arXiv preprint arXiv:2307.09288}.

\bibitem[{Tunstall et~al.(2023)Tunstall, Beeching, Lambert, Rajani, Rasul, Belkada, Huang, von Werra, Fourrier, Habib et~al.}]{zephyr}
Lewis Tunstall, Edward Beeching, Nathan Lambert, Nazneen Rajani, Kashif Rasul, Younes Belkada, Shengyi Huang, Leandro von Werra, Cl{\'e}mentine Fourrier, Nathan Habib, et~al. 2023.
\newblock Zephyr: Direct distillation of lm alignment.
\newblock \emph{Conference on Language Modeling}.

\bibitem[{Wang et~al.(2024{\natexlab{a}})Wang, Jiang, Yang, Liu, and Chen}]{a_kl_dpo}
Chaoqi Wang, Yibo Jiang, Chenghao Yang, Han Liu, and Yuxin Chen. 2024{\natexlab{a}}.
\newblock Beyond reverse kl: Generalizing direct preference optimization with diverse divergence constraints.
\newblock In \emph{The Twelfth International Conference on Learning Representations}.

\bibitem[{Wang et~al.(2023)Wang, Li, Chen, Cai, Zhu, Lin, Cao, Liu, Liu, and Sui}]{llm_as_judge1}
Peiyi Wang, Lei Li, Liang Chen, Zefan Cai, Dawei Zhu, Binghuai Lin, Yunbo Cao, Qi~Liu, Tianyu Liu, and Zhifang Sui. 2023.
\newblock Large language models are not fair evaluators.
\newblock In \emph{the 62nd Annual Meeting of the Association for Computational Linguistics (Volume 1: Long Papers).}

\bibitem[{Wang et~al.(2024{\natexlab{b}})Wang, Zhang, Zhao, Tan, and Cam-Tu}]{emnlp_wsq}
Shiqi Wang, Zhengze Zhang, Rui Zhao, Fei Tan, and Nguyen Cam-Tu. 2024{\natexlab{b}}.
\newblock Reward difference optimization for sample reweighting in offline {RLHF}.
\newblock In \emph{Findings of the Association for Computational Linguistics: EMNLP 2024}.

\bibitem[{Xu et~al.(2023)Xu, Guo, Duan, and McAuley}]{baize}
Canwen Xu, Daya Guo, Nan Duan, and Julian McAuley. 2023.
\newblock Baize: An open-source chat model with parameter-efficient tuning on self-chat data.
\newblock In \emph{Proceedings of the 2023 Conference on Empirical Methods in Natural Language Processing}, pages 6268--6278.

\bibitem[{Yang et~al.(2024)Yang, Yang, Zhang, Hui, Zheng, Yu, Li, Liu, Huang, Wei et~al.}]{qwen2.5}
An~Yang, Baosong Yang, Beichen Zhang, Binyuan Hui, Bo~Zheng, Bowen Yu, Chengyuan Li, Dayiheng Liu, Fei Huang, Haoran Wei, et~al. 2024.
\newblock Qwen2. 5 technical report.
\newblock \emph{arXiv preprint arXiv:2412.15115}.

\bibitem[{Zeng et~al.(2024)Zeng, Liu, Ma, Yang, Zhang, and Wang}]{token_dpo}
Yongcheng Zeng, Guoqing Liu, Weiyu Ma, Ning Yang, Haifeng Zhang, and Jun Wang. 2024.
\newblock Token-level direct preference optimization.
\newblock \emph{Proceedings of the 41st International Conference on Machine Learning}.

\bibitem[{Zhang et~al.(2024)Zhang, Wu, Li, Yang, Zhao, Jiang, and Tan}]{tanfei1_sft}
Hengyuan Zhang, Yanru Wu, Dawei Li, Sak Yang, Rui Zhao, Yong Jiang, and Fei Tan. 2024.
\newblock Balancing speciality and versatility: a coarse to fine framework for supervised fine-tuning large language model.
\newblock In \emph{Findings of the Association for Computational Linguistics ACL 2024}, pages 7467--7509.

\bibitem[{Zhao et~al.(2023)Zhao, Joshi, Liu, Khalman, Saleh, and Liu}]{SLiC-HF2023}
Yao Zhao, Rishabh Joshi, Tianqi Liu, Misha Khalman, Mohammad Saleh, and Peter~J Liu. 2023.
\newblock Slic-hf: Sequence likelihood calibration with human feedback.
\newblock \emph{Advances in Neural Information Processing Systems}.

\bibitem[{Zheng et~al.(2023)Zheng, Chiang, Sheng, Zhuang, Wu, Zhuang, Lin, Li, Li, Xing et~al.}]{vicuna}
Lianmin Zheng, Wei-Lin Chiang, Ying Sheng, Siyuan Zhuang, Zhanghao Wu, Yonghao Zhuang, Zi~Lin, Zhuohan Li, Dacheng Li, Eric Xing, et~al. 2023.
\newblock Judging llm-as-a-judge with mt-bench and chatbot arena.
\newblock \emph{Advances in Neural Information Processing Systems}, 36:46595--46623.

\end{thebibliography}
\appendix
\newpage
\onecolumn
\section{Proof of Theorem 1}\label{sec:appendix}
Assume that $\beta_1+\beta_2>0,\beta_1\geq0,\beta_2\geq 0$, we have:

\begin{align}
    &\arg_{\theta}\max{\mathbb{E}_{x\sim \mathcal{D},y\sim \pi_\theta(x)}\left[r(x,y)\right]-\beta_1D_{KL}[\pi_\theta(y|x)||\pi_{ref}(y|x)]-\beta_2D_{KL}[\pi_\theta(y|x)||\pi_{te}(y|x)]}\nonumber\\
    &=\arg_{\theta}\max{\mathbb{E}_{x\sim \mathcal{D}}\mathbb{E}_{y\sim \pi_\theta(x)}\left[r(x,y)-\beta_1\log\frac{\pi_\theta(y|x)}{\pi_{ref}(y|x)}-\log\beta_2\frac{\pi_\theta(y|x)}{\pi_{te}(y|x)}\right]}\nonumber\\
    &=\arg_{\theta}\min{\mathbb{E}_{x\sim \mathcal{D}}\mathbb{E}_{y\sim \pi_\theta(x)}\left[(\beta_1+\beta_2)\log\frac{\pi_\theta(y|x)}{\pi_{ref}(y|x)^\frac{\beta_1}{\beta_1+\beta_2}*\pi_{te}(y|x)^{\frac{\beta_2}{\beta_1+\beta_2}}}-r(x,y)\right]}\nonumber\\
    &=\arg_{\theta}\min{\mathbb{E}_{x\sim \mathcal{D}}\mathbb{E}_{y\sim \pi_\theta(x)}\left[\log \frac{\pi_\theta(y|x)}{\pi_{ref}(y|x)^\frac{\beta_1}{\beta_1+\beta_2}*\pi_{te}(y|x)^{\frac{\beta_2}{\beta_1+\beta_2}}\exp(\frac{1}{\beta_1+\beta_2}r(x,y))}\right]}\nonumber\\
    &=\arg_{\theta}\min{\mathbb{E}_{x\sim \mathcal{D}}\mathbb{E}_{y\sim \pi_\theta(x)}\left[ \log\frac{\pi_\theta(y|x)}{\frac{1}{Z(x)}\pi_{ref}(y|x)^\frac{\beta_1}{\beta_1+\beta_2}*\pi_{te}(y|x)^{\frac{\beta_2}{\beta_1+\beta_2}}\exp(\frac{1}{\beta_1+\beta_2}r(x,y))}-\log Z(x)\right]} \label{proof-loss}
\end{align}

where $Z(x)=\sum_y\pi_{ref}(y|x)^\frac{\beta_1}{\beta_1+\beta_2}*\pi_{te}(y|x)^{\frac{\beta_2}{\beta_1+\beta_2}}\exp(\frac{1}{\beta_1+\beta_2}r(x,y))$

Note that the partition function is a function of only $x$ and the reference policy $\pi_{ref}$ and the teacher policy $\pi_{te}$, but does not depend on the policy $\pi_\theta$. We can now define 
\begin{align}
    \pi^*(y|x)=\frac{1}{Z(x)}\pi_{ref}(y|x)^\frac{\beta_1}{\beta_1+\beta_2}*\pi_{te}(y|x)^{\frac{\beta_2}{\beta_1+\beta_2}}\exp(\frac{1}{\beta_1+\beta_2}r(x,y))
\end{align}

which is a valid distribution as $\pi^*(y|x)>0$  for all $y$ and $\sum_{y}\pi^*(y|x)=1$. For $x\sim\mathcal{D},y\sim\pi_\theta(x)$We can then re-organize the objective in eq.\ref{proof-loss} as:
\begin{align}
&\max\mathop{\mathbb{E}}\limits_{x,y}\left[r(x,y)\right]-\beta_1D_{KL}[\pi_\theta(y|x)||\pi_{ref}(y|x)]-\beta_2D_{KL}[\pi_\theta(y|x)||\pi_{te}(y|x)]\nonumber\\
    &=\min{\mathbb{E}_{x\sim \mathcal{D}}\left[\mathbb{E}_{y\sim \pi_\theta(x)}\left[\log\frac{\pi_\theta(y|x)}{\pi^*(y|x)}\right]-\log Z(x)\right]}\nonumber\\
    &=\min{\mathbb{E}_{x\sim \mathcal{D}}\left[D_{KL}[\pi_\theta(y|x)||\pi^*(y|x)]-\log Z(x)\right]}
\end{align}
Because $\log Z(x)$ is independent of $\pi_\theta$ thus the minimum is only decided by the KL term. According to Gibbs' inequality, the KL-divergence is minimized at 0 if and only if the two distributions are identical. Hence we have proven that the optimal policy of the original objective in eq.\ref{original_loss} is 
\begin{align}
\pi^*(y|x)=\frac{1}{Z(x)}\pi_{ref}(y|x)^\frac{\beta_1}{\beta_1+\beta_2}\pi_{te}(y|x)^{\frac{\beta_2}{\beta_1+\beta_2}}\nonumber *\exp(\frac{1}{\beta_1+\beta_2}r(x,y))
\end{align}
According to theorem 1, with simple algebra, we can derive the following relationship between the reward model and $\pi^*$
\begin{align}
    r(x,y)=(\beta_1+\beta_2)\frac{\log(\pi_\theta^*(y|x))}{\log{}\pi_{ref}(y|x)^{\frac{\beta_1}{\beta_1+\beta_2}}\pi_{te}(y|x)^{\frac{\beta_2}{\beta_1+\beta_2}}} +(\beta_1+\beta_2) \log Z(x)
\end{align}
Similar to DPO \cite{dpo}, because $\log Z(x)$ is not related to $y$ and thus under the Bradley-Terry model, we will get preference probability in terms of only the optimal policy $\pi^*$, reference policy $\pi_{ref}$ and teacher LLM $\pi_{te}$.
\begin{equation}
    p^*(y_1\succ y_2)= \frac{1}{1+\exp(\frac{(\beta_1+\beta_2)\log(\pi_\theta^*(y_2|x))}{\log{\pi_{ref}(y_2|x)^{\beta_1/(\beta_1+\beta_2)}\pi_{te}(y_2|x)^{\beta_2/(\beta_1+\beta_2)}}}-\frac{(\beta_1+\beta_2)\log(\pi_\theta^*(y_1|x))}{\log{\pi_{ref}(y_1|x)^{\beta_1/(\beta_1+\beta_2)}\pi_{te}(y_1|x)^{\beta_2/(\beta_1+\beta_2)}}})}\label{eq:long-p}
\end{equation}
\twocolumn

\section{Future Works}
Several promising directions for future research emerge from this work:
\begin{enumerate}
\item \textbf{Extended Experimentation on Reasoning Tasks.} Recent developments with DeepSeek-R1 \cite{deepseekr1} have demonstrated remarkable capabilities in reasoning and coding tasks, with the model's open-source nature enabling distillation experiments. Evidence suggests that single dSFT distillation significantly enhances the Qwen model's reasoning capabilities. Further investigation is warranted to examine whether dDPO and \textit{daDPO} can provide additional improvements in this context.

\item \textbf{Integration with Alternative Preference Learning Frameworks.} While our method has successfully incorporated teacher distribution information into the original DPO framework, additional theoretical analysis could illuminate potential integrations with DPO variants, such as KTO \cite{ethayarajh2024kto} and f-DPO \cite{a_kl_dpo}.

\item \textbf{Expansion to Diverse Teacher LLMs.} Our current experiments primarily utilize a single-teacher LLM. Future work should explore scenarios involving multiple teacher responses from diverse sources, similar to Zephyr \cite{zephyr}, to better understand the impact of teacher diversity and ensemble effects.
\end{enumerate}
\section{Algorithm}
 The algorithm of our proposed \textit{daDPO} is shown in Algorithm \ref{algo}. This demonstrates that our algorithm is easy to implement and requires only a few additional lines of code compared to DPO.  The algorithm contains three steps: Firstly, we leverage datasets with prompts only and then sample teachers' and students' responses to them. After that, we first run dSFT on datasets with the teacher's responses as golden answers. Finally, we run \textit{daDPO} training with our proposed loss in eq.\ref{daDPO_loss}.

\begin{algorithm*}[htbp]
\caption{distribution-aware Direct Preference Optimization (daDPO) for distillation}
\begin{algorithmic}[1]
\State \textbf{Input:}$ \mathcal{C}_p = \{x_i\}^N$,$\pi_{st}$, $\pi_{te}$, $\beta_1,\beta_2$ \Comment{Stand for Prompt dataset, student model, teacher model, hyper-parameters respectively}
\State \textbf{Output:} Optimized policy model $\pi_\theta$
\State Sample teachers responses $\{y^{t}\}^N$ from $\pi_{te}$ on $\mathcal{C}_p$
\State Sample student responses $\{y^{s}\}^N$  from $\pi_{st}$ on $\mathcal{C}_p$
\State Get the preference dataset $\mathcal{C}_{dpo}=\{(x,y^t,y^s)\}^N$ and sft dataset $\mathcal{C}_{sft}=\{(x,y^t)\}^N$
\State $\pi_{sft}\leftarrow$  Optimize $\pi_{st}$ on $\mathcal{C}_{sft}$ with sft loss in eq.\ref{eq:sft-loss}
\State $\pi_{ref} \leftarrow\pi_{sft},\pi_\theta\leftarrow\pi_{sft}$
\For{each training batch}
    \State Sample batch $(x, y^t, y^s) \sim \mathcal{C}_{dpo}$
    \State $r^w = \log \pi_\theta(y^t|x) - \log \pi_{ref}(y^t|x)$ \Comment{Log ratio for winning response}
    \State $r^l = \log \pi_\theta(y^s|x) - \log \pi_{ref}(y^s|x)$ \Comment{Log ratio for losing response}
    \State $r^w_t = \log \pi_\theta(y^t|x) - \log \pi_{te}(y^t|x)$ \Comment{Log ratio for winning response with teacher model as reference}
    \State $r^l_t = \log \pi_\theta(y^s|x) - \log \pi_{te}(y^s|x)$ \Comment{Log ratio for losing response with teacher model as reference}
    \State Compute loss:
    \State $\mathcal{L}_{daDPO} = -\mathbb{E}_{(x,y^w,y^l)\sim\mathcal{D}}[\log\sigma(\beta_1(r^w - r^l)+\beta_2(r^w_t-r^l_t))]$
    \State Update parameters $\theta$ to minimize $\mathcal{L}_{daDPO}$
\EndFor
\State \textbf{Return} Distilled model $\pi^*_\theta$
\end{algorithmic}
\label{algo}
\end{algorithm*}
\section{Details for baselines}\label{sec:appendix_baseline}

\torevise{We leverage the same decoding strategy (i.e., generation config) and avoid much tuning of it for all methods with greedy decoding (i.e., sampling=False). Training details are as follows:} % delete the lr and bsz configuration
\begin{enumerate}
    \item dSFT \cite{zephyr}: We use \(\{(x, y^t)\}\) for Supervised Fine-Tuning of $\pi_{st}$. 
    \item dSFT+KL \cite{prune_kd,minillm_kl}: We use \(\{(x, y^t)\}\) to supervise and fine-tune the student model, and incorporate a token-level KL divergence loss defined in Eq. \ref{kl-loss}. The weight of KL divergence loss is tuned within the range \([0.1,0.2,0.4]\). 
    \item dDPO \cite{zephyr,qdpo,smallbutfunny}: We use  \(\{(x, y^t, y^s)\}\) to optimize the student model after dSFT by DPO algorithm. We tune \(\beta\) of DPO within the range \([0.01, 0.1, 1.0]\).  
    \item dDPO+KL: We use  \(\{(x, y^t, y^s)\}\) to optimize the student model after dSFT, and then add a token-level KL divergence loss defined in Eq. \ref{kl-loss} into DPO loss. For hyperparameters, we tune the weight of KL divergence loss within the range \([0.1,0.2,0.4]\).
\end{enumerate}
\begin{figure}[t]
\centering
  \includegraphics[width=0.98\linewidth]{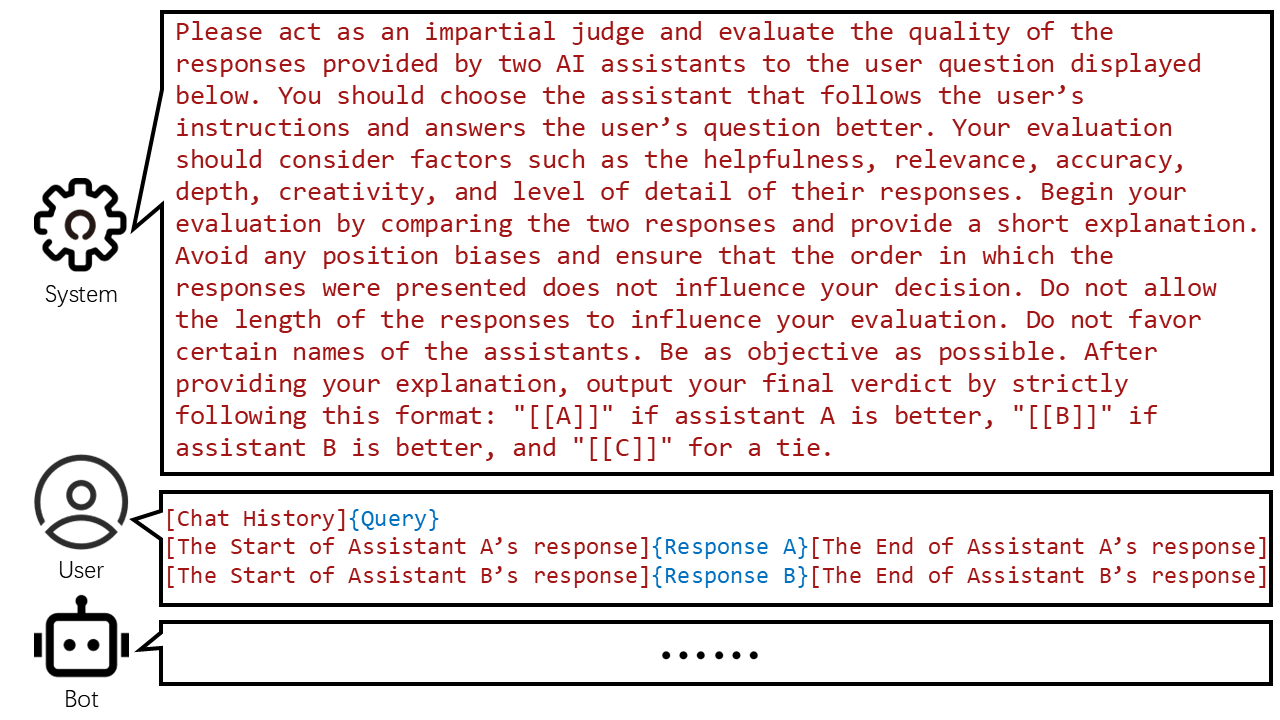}
  \caption {Evaluation prompt for GPT-4o-2024-08-06.}
  \label{fig:prompt}
\end{figure}
\section{Supplementary experiments}

\subsection{Evaluation on More Challenging Benchmarks}
\torevise{
Recently, several more challenging benchmarks have been introduced to more accurately assess the capabilities of modern large language models (LLMs), such as Arena-Hard and AlpacaEval 2.0. However, these benchmarks are excessively difficult for weaker models like those in the Vicuna Series. For instance, the Vicuna-7B model achieves only 7.2\% on AlpacaEval 2.0\footnote{\url{https://tatsu-lab.github.io/alpaca_eval/}}, and even the larger Vicuna-33B model attains only 8.6\% on Arena-Hard\footnote{\url{https://github.com/lmarena/arena-hard-auto?tab=readme-ov-file\#leaderboard}}.}

\torevise{In this subsection, we present the performance of our Qwen Series distillation experiments on these challenging benchmarks, where the teacher model, Qwen2.5-7B, demonstrates reasonably strong results. The results are shown in table \ref{tab:chanllenge} which makes our work more convincing and sound.}

\subsection{Comparison with the existing RLHF alternative}
\torevise{
We compare with one more offline RLHF method for distillation: dKTO.  We use  \(\{(x, y^t, y^s)\}\) to optimize the student model after dSFT by KTO algorithm. We tune \(\beta\) of KTO \cite{ethayarajh2024kto} within the range of [0.01, 0.1, 0.5, 1.0, 10.0]. 
Results are shown in table \ref{tab:kto}.
}
\begin{table}[t]
\centering
\begin{tabular}{@{}l|c|c@{}}
\toprule
\textbf{Method}                            & \textbf{MT-Bench}  & \textbf{Indomain} \\ \midrule
Vicuna-7B                                   & 4.74                                & 0.0\%                   \\
\midrule
Vicuna-7B\textsubscript{0.8}                            & 3.06                          & -50.6\%                    \\
\quad +dSFT                             & 3.68                                     & -40.0\%                      \\
\quad +dDPO                             & 3.78                               & -31.0\%                      \\
\quad +dKTO                             & 3.75                               & -19.0\%                      \\
\rowcolor{light-gray}\quad+daDPO                             &    3.84                                               &-7.3\%                     

\end{tabular}

\caption{Performance of daDPO on pruned Vicuna-7B compared to different baselines without distribution including  RLHF alternatives. Here, Vicuna-7B\textsubscript{0.8} refer to the model with 20\%  of the parameters pruned.}

\label{tab:kto}
\end{table}
\begin{table}[htbp]

\begin{tabular}{@{}l|c|c@{}}
\toprule
\textbf{Method}                            & \textbf{Arena-Hard}  & \textbf{AlpacaEval-2.0} \\ \midrule
Qwen2.5-7B	   & 54.1\%	                              & 28.34\%                    \\
\midrule
Qwen2.5-1.5B	                           & 10.7\%	                & 6.44\%                   \\
\quad +dDPO	                      &21.5\%	                   & 14.66\%                        \\
\rowcolor{light-gray}\quad +daDPO                           & 	22.4\%	                 & 16.41\%                    
\end{tabular}
\caption{Performance of daDPO on Qwen2.5 Series models on more challenging benchmarks. }
\label{tab:chanllenge}
\end{table}

\section{LLM-as-judge evaluation details}
We use "GPT-4o-2024-08-06" for In-domain evaluation. Following some previous studies \cite{vicuna,llm_as_judge1,tanfei4_segmentation,tanfei3_pretain}, we ask the LLMs to determine which of the two responses under the same query is better. The prompt for evaluation is as Figure~\ref{fig:prompt}
\subsection{In-domain evaluation results details}
This section presents the details of the in-domain evaluation results in section 5, including Vicuna-7B, Qwen2.5 Series, and LLama3.2-1B. The detailed results are shown in Tables \ref{tab:detail1}, \ref{tab:detail2}, and \ref{tab:detail3}.
\begin{table*}[t]
\resizebox{\textwidth}{!}{%
\begin{tabular}{cll|ccc|ccc|ccc|ccc|ccc}
\hline
\multicolumn{3}{c|}{\textit{}} & \multicolumn{3}{c|}{Original} & \multicolumn{3}{c|}{+dSFT} & \multicolumn{3}{c|}{+dDPO} & \multicolumn{3}{c|}{+rDPO}& \multicolumn{3}{c}{+daDPO} \\ \hline
\multicolumn{3}{c|}{Model} &  Win    & Tie    &  Lose   &  Win    & Tie    & Lose  & Wine   & Tie    & Lose   & Win    & Tie    & Lose  & Win    & Tie    & Lose  \\ \midrule
\multicolumn{3}{c|}{Vicuna-7B}    & 0 & 300 & 0 & - & - & - & - & - & - & -& - & - & -& - & - \\
\multicolumn{3}{c|}{Vicuna-7B\textsubscript{0.8}}   & 62 & 24 & 214 & 71 & 38  & 191 & 82 & 43  & 175 & 117 & 39  & 144  & 120 & 38  & 142 \\
\multicolumn{3}{c|}{Vicuna-7B\textsubscript{0.5}}    & 15 & 10 & 275 & 17 & 10  & 262 & 27 & 25  & 248 & 23 & 23  & 254  & 37 & 22  & 241\\
 \hline
\end{tabular}%
}
\caption{Details of In-domain Evaluation Versus Vicuna-7B in Table \ref{tab:pruned-result}.}
\label{tab:detail1}
\end{table*}
\begin{table*}[htbp]
\resizebox{\textwidth}{!}{%
\begin{tabular}{cll|ccc|ccc|ccc|ccc|ccc}
\hline
\multicolumn{3}{c|}{\textit{}} & \multicolumn{3}{c|}{Original} & \multicolumn{3}{c|}{+dSFT} & \multicolumn{3}{c|}{+dDPO} & \multicolumn{3}{c|}{+rDPO}& \multicolumn{3}{c}{+daDPO} \\ \hline
\multicolumn{3}{c|}{Model} &  Win    & Tie    &  Lose   &  Win    & Tie    & Lose  & Wine   & Tie    & Lose   & Win    & Tie    & Lose  & Win    & Tie    & Lose  \\ \midrule
\multicolumn{3}{c|}{Qwen2.5-7B}    & 0 & 300 & 0 & - & - & - & - & - & - & -& - & - & -& - & - \\
\multicolumn{3}{c|}{Qwen2.5-1.5B}   & 41 & 33 & 226 & 124 & 30  & 146 & 138 & 26  & 136 & 130 & 25  & 145  & 161 & 20  & 119 \\
\multicolumn{3}{c|}{Qwen2.5-0.5B}    & 29 & 17 & 254 & 43 & 15  & 242 & 47 & 16  & 237 & 68 & 18  & 214  & 75 & 16 & 209\\
 \hline
\end{tabular}%
}
\caption{Details of In-domain Evaluation Versus Qwen2.5-7B in Table \ref{tab:qwen-series-results}.}
\label{tab:detail2}
\end{table*}
\begin{table*}[htbp]
\resizebox{\textwidth}{!}{%
\begin{tabular}{cll|ccc|ccc|ccc|ccc|ccc}
\hline
\multicolumn{3}{c|}{\textit{}} & \multicolumn{3}{c|}{Original} & \multicolumn{3}{c|}{+dSFT} & \multicolumn{3}{c|}{+dDPO} & \multicolumn{3}{c|}{+rDPO}& \multicolumn{3}{c}{+daDPO} \\ \hline
\multicolumn{3}{c|}{Model} &  Win    & Tie    &  Lose   &  Win    & Tie    & Lose  & Wine   & Tie    & Lose   & Win    & Tie    & Lose  & Win    & Tie    & Lose  \\ \midrule
\multicolumn{3}{c|}{Qwen2.5-7B}    & 0 & 300 & 0 & - & - & - & - & - & - & -& - & - & -& - & - \\
\multicolumn{3}{c|}{LLama3.2-1B}   & 38 & 7 & 255& 79 & 28  & 193 & 81 & 29  & 190 & 76 & 30  & 194  & 82 & 28  & 190 \\
 \hline
\end{tabular}%
}
\caption{Details of In-domain Evaluation Versus Qwen2.5-7B in Table \ref{tab:different-series-results}.}
\label{tab:detail3}
\end{table*}

\section{Case Study}
We show cases in Table \ref{tab:case-study1} and Table \ref{tab:case-study2} in this section to compare responses on the MTbench of dDPO and \textit{daDPO}. The student model is Vicuna-7B-20\% pruned model and the teacher model is Vicuna-7B. These cases show that integrating distribution information into DPO helps the distilled model better understand the instruction and thus output better responses. 
\begin{table*}[t]
    \centering
    \begin{tabular}{p{16cm}}
    \toprule
        \textbf{Round 1} \textbf{Human:} Provide insights into the correlation between economic indicators such as GDP, inflation, and unemployment rates. Explain how fiscal and monetary policies affect those indicators.?
        \paragraph{+daDPO} Economic indicators such as GDP, inflation, and unemployment rates are closely correlated as they provide insights into the health and performance of an economy.GDP (Gross Domestic Product) is the total value of all goods and services produced within a country over a specific period of time...
        \paragraph{+dDPO} Economic indicators such as GDP, inflation, and unemployment rates are closely interconnected and can influence each other in a complex manner. Here are some insights into the correlation between economic indicators and their impact on fiscal and monetary policies: 1. GDP: GDP stands for gross domestic product, which is...
        \\\midrule
        \textbf{Round 2} \textbf{Human: }Now, \textbf{\textit{explain them again like I'm five}}
        
        \paragraph{+daDPO} Economic indicators such as GDP, inflation, and unemployment rates tell us how well our economy is doing. \textbf{GDP is like a report card} for how much our country is making from producing and selling things. Inflation is like a report card for how much things are getting more expensive or cheaper. Unemployment rate is like a report card for how many people are having trouble finding a job. \textbf{Fiscal policies are like the rules our government makes to help or hurt our economy}... \textcolor{blue}{Rating: [[8]]}
        \paragraph{+dDPO} Sure! Let's start with the first important factor, GDP (Gross Domestic Product): GDP stands for Gross Domestic Product, which is the total output of a country's economy. It's a measure of the country's economic health and well-being, and it's important because it reflects the production and consumption of goods and services in the economy...\textcolor{red}{(Still too professional for a kid)} \textcolor{blue}{Rating: [[4]]}
     \\
        \bottomrule
    \end{tabular}
    \caption{Case Study on MT-bench: Responses comparison between dDPO and daDPO; Integrating distribution information into DPO helps on instruction following ability.}
    \label{tab:case-study1}
\end{table*}
\begin{table*}[t]
    \centering
    \begin{tabular}{p{16cm}}
    \toprule
        \textbf{Round 1} \textbf{Human:} Write a persuasive email to convince your introverted friend, who dislikes public speaking, to volunteer as a guest speaker at a local event. Use compelling arguments and address potential objections. Please be concise.
        \paragraph{+daDPO} Subject: Convince Your Introverted Friend to Volunteer as a Guest Speaker at a Local Event \textbackslash n\textbackslash nDear [Friend's Name],\textbackslash n\textbackslash nI hope this email finds you well. I know how much you dislike public speaking, but I also know how much your presence and insights can make a difference at a local event. That's why I'm reaching out to convince you to volunteer as a guest speaker at this upcoming event. Here are some compelling arguments to consider:...
        \paragraph{+dDPO} Subject: Encouraging Your Introverted Friend to Be a Speaker at a Local Event \textbackslash n\textbackslash nDear [Friend's Name],\textbackslash n\textbackslash n I hope this email finds you well. I wanted to reach out and see if you're open to considering being a speaker at one of our local events. I believe that you have a unique perspective and voice that could add a lot to the conversation, and I think that being a part of this event would be a great way for you to connect with others and share your ideas...
        \\ \midrule
        \textbf{Round 2} \textbf{Human: }Can you rephrase your previous answer and \textbf{\textit{incorporate a metaphor or simile in each sentence}}? 
        
        \paragraph{+daDPO} Subject: Convince Your Introverted Friend to Volunteer as a Guest Speaker at a Local Event with Metaphor and Simile \textbackslash n\textbackslash n
        Dear [Friend's Name], \textbackslash n\textbackslash n I hope this email finds you well. I know how much you dislike public speaking, but \textbf{think of it as climbing a mountain} - it can be intimidating, but the view from the top is worth it. By volunteering as a guest speaker at this local event, you have the opportunity to make a positive impact and grow personally, \textbf{like polishing a diamond with constant refinement}. Here are some compelling arguments to consider, along with a metaphor and simile:... \textcolor{blue}{Rating: [[8]]}
        \paragraph{+dDPO} Sure, here's a revised version of the email that incorporates metaphors and similes:\textbackslash n\textbackslash nSubject: Encouraging Your Introverted Friend to Be a Speaker at a Local Event\textbackslash n\textbackslash nDear [Friend's Name],\textbackslash n\textbackslash nI hope this email finds you well. I wanted to reach out and see if you're open to considering being a speaker at one of our local events. I believe that you have a unique perspective and a valuable voice, and I'm excited to see what you can do to contribute to this event.\textbackslash n\textbackslash nAs you may know, public speaking can be a daunting task, especially for introverts who may feel more comfortable being a part of a group. However, I want to assure you that there are many ways to overcome this fear and make the most of your public speaking experience.\textbackslash n\textbackslash n One way to prepare for a speech is to practice, practice, and surround yourself with supportive people. It's important to have a plan in place and to be confident in your ability to deliver a great speech... \textcolor{red}{(No metaphor or simile  in this response)} \textcolor{blue}{Rating: [[3]]}
     \\
        \bottomrule
    \end{tabular}
    \caption{Case Study on MT-bench: Responses comparison between dDPO and daDPO; Integrating distribution information into DPO helps on instruction following ability.}
    \label{tab:case-study2}
\end{table*}

% Please add the following required packages to your document preamble:
% \usepackage{graphicx}
% Please add the following required packages to your document preamble:
% \usepackage{graphicx}

\end{document}